\pdfoutput=1

\documentclass[11pt]{article}
\usepackage{float}

\usepackage{acl}

\usepackage[utf8]{inputenc}
\usepackage{pgfplots}
\usepackage{dsfont}

\DeclareUnicodeCharacter{2212}{−}
\usepgfplotslibrary{groupplots,dateplot}
\usetikzlibrary{patterns,shapes.arrows}
\pgfplotsset{compat=newest}

\usepackage{tikzscale}
\usepackage{amsmath}

\usepackage{multirow, colortbl}
\usepackage{color}
\usepackage{array}
\usepackage{multirow}
\usepackage{CJKutf8}
\usepackage{tabularx,booktabs}
\usepgfplotslibrary{groupplots}
\usepackage{makecell}
\usepackage{enumitem}
\usepackage{placeins}

\usepackage[normalem]{ulem}
\usepackage{graphicx}
\usepackage{booktabs}

\usepackage{pifont}

\usepackage{pgfplots}
\pgfplotsset{width=10cm,compat=1.9}

\definecolor{ablation6}{HTML}{fcefed}
\definecolor{ablation_tie}{HTML}{fce3e1}

\definecolor{ablation5}{HTML}{fcd8d4}
\definecolor{ablation4}{HTML}{FBC3BC}
\definecolor{ablation3}{HTML}{F7A399}
\definecolor{ablation2}{HTML}{F38375}
\definecolor{ablation1}{HTML}{EF6351}

\usepackage[]{algpseudocode}
\usepackage[]{algorithm}
\usepackage{float}
\algtext*{EndFor}%
\algtext*{EndProcedure}%

\usepackage{booktabs}
\usepackage{times}
\usepackage{latexsym}
\usepackage{adjustbox}

\usepackage[T1]{fontenc}
\usepackage[utf8]{inputenc}
\usepackage[russian,english]{babel}
\newcommand{\russian}[1]{{\fontencoding{T2A}\selectfont\foreignlanguage{russian}{#1}}}

\usepackage{amsmath}
\usepackage{amssymb}
\usepackage{booktabs}
\usepackage{multirow}
\usepackage{scalerel,xparse}
\usepackage[utf8]{inputenc}
\usepackage{cleveref}
\usepackage{microtype}
\usepackage{enumitem}\usepackage[utf8]{inputenc}
\usepackage{pgfplots}
\usepackage{dsfont}

\DeclareUnicodeCharacter{2212}{−}
\usepgfplotslibrary{groupplots,dateplot}
\usetikzlibrary{patterns,shapes.arrows}
\pgfplotsset{compat=newest}

\usepackage{tikzscale}
\usepackage{amsmath}
\newcommand{\probP}{\text{I\kern-0.15em P}}

\usepackage{multirow, colortbl}
\usepackage{color}
\usepackage{array}
\usepackage{multirow}
\usepackage{CJKutf8}
\usepackage{tabularx,booktabs}
\usepgfplotslibrary{groupplots}
\usepackage{makecell}
\usepackage{enumitem}
\usepackage{placeins}

\usepackage{soul}
\definecolor{light blue}{RGB}{215, 242, 252}
\definecolor{light purple}{RGB}{247, 215, 252}
\definecolor{light orange}{rgb}{0.9961, 0.875, 0.7188}
\sethlcolor{light blue}
\newcommand{\hlpurple}[1]{\sethlcolor{light purple}\hl{#1}\sethlcolor{light blue}}
\newcommand{\hlorange}[1]{\sethlcolor{light orange}\hl{#1}\sethlcolor{light blue}}

\usepackage{graphicx}
\usepackage{booktabs}

\usepackage{pgfplots}
\pgfplotsset{width=10cm,compat=1.9}

\definecolor{ablation6}{HTML}{fcefed}
\definecolor{ablation_tie}{HTML}{fce3e1}

\definecolor{ablation5}{HTML}{fcd8d4}
\definecolor{ablation4}{HTML}{FBC3BC}
\definecolor{ablation3}{HTML}{F7A399}
\definecolor{ablation2}{HTML}{F38375}
\definecolor{ablation1}{HTML}{EF6351}

\usepackage[]{algpseudocode}
\usepackage[]{algorithm}
\usepackage{float}
\algtext*{EndFor}%
\algtext*{EndProcedure}%

\usepackage{times}
\usepackage{latexsym}
\usepackage{adjustbox}

\usepackage[T1]{fontenc}
\usepackage[utf8]{inputenc}
\usepackage[russian,english]{babel}

\usepackage{amsmath}
\usepackage{amssymb}
\usepackage{booktabs}
\usepackage{multirow}
\usepackage{scalerel,xparse}
\usepackage[utf8]{inputenc}
\usepackage{cleveref}
\usepackage{microtype}
\usepackage{enumitem}
\usepackage{placeins}

\crefformat{section}{\S#2#1#3}
\crefformat{subsection}{\S#2#1#3}
\crefformat{subsubsection}{\S#2#1#3}

\definecolor{zoey green}{rgb}{0.684,0.836,0.227}

\newcommand{\ignore}[1]{}

\title{Automatic Input Rewriting Improves Translation \\ with Large Language Models}

\author{Dayeon Ki \\
  University of Maryland \\
  \texttt{dayeonki@umd.edu} \\\And
  Marine Carpuat \\
  University of Maryland \\
  \texttt{marine@cs.umd.edu} \\}

\begin{document}
\maketitle


\begin{abstract}
Can we improve machine translation (MT) with LLMs by rewriting their inputs automatically? Users commonly rely on the intuition that well-written text is easier to translate when using off-the-shelf MT systems. LLMs can rewrite text in many ways but in the context of MT, these capabilities have been primarily exploited to rewrite outputs via post-editing. We present an empirical study of 21 input rewriting methods with 3 open-weight LLMs for translating from English into 6 target languages. We show that text simplification is the most effective MT-agnostic rewrite strategy and that it can be improved further when using quality estimation to assess translatability. Human evaluation further confirms that simplified rewrites and their MT outputs both largely preserve the original meaning of the source and MT. These results suggest LLM-assisted input rewriting as a promising direction for improving translations.\footnote{We release our code and dataset at \url{https://github.com/dayeonki/rewrite_mt}.}

\end{abstract}

\section{Introduction}

\definecolor{light blue}{RGB}{215, 242, 252}
\definecolor{light purple}{RGB}{247, 215, 252}
\definecolor{light orange}{rgb}{0.9961, 0.875, 0.7188}

\begin{figure*}
    \centering
    \includegraphics[width=\textwidth]{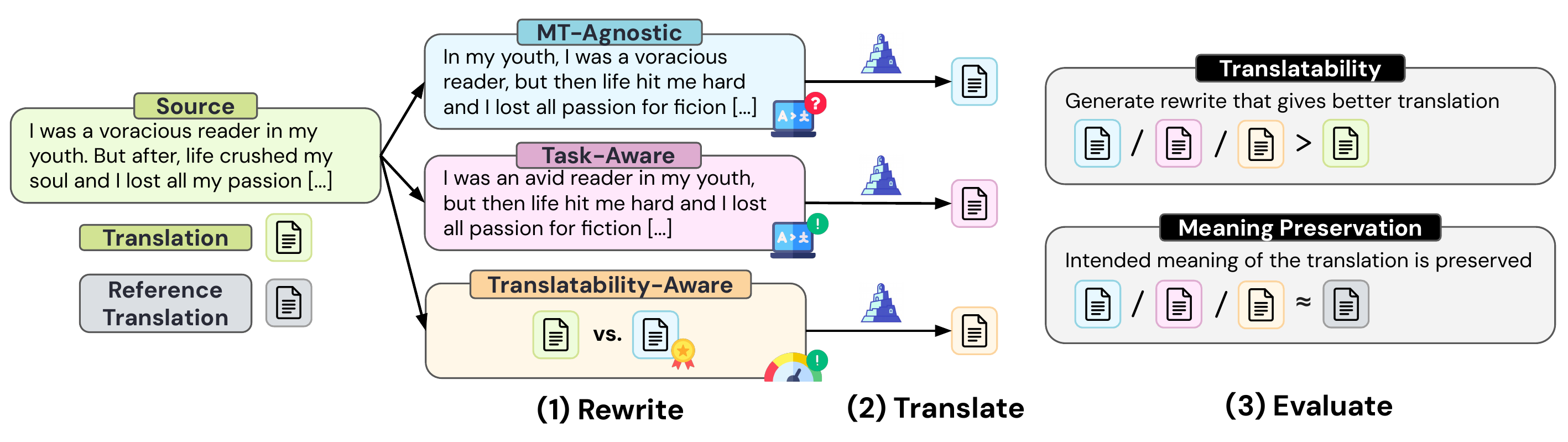}
    \caption{Overview of the rewriting pipeline. \textbf{(1) Rewrite:} Given source sentence, we generate rewrites using different rewriting methods: \hl{MT Agnostic}, \hlpurple{Task-Aware} and \hlorange{Translatability-Aware}. \textbf{(2) Translate:} We translate each rewrite using our MT system, \textsc{Tower-Instruct 7B}. \textbf{(3) Evaluate:} We automatically evaluate rewrites along translatability, meaning preservation, and overall translation quality.}
    \label{fig:main_figure}
\end{figure*}

Machine translation (MT) users and developers have long exploited the idea that some texts are easier to translate than others. For instance, guiding people to edit their inputs so that they are well formed is a cornerstone of MT literacy courses \citep{bowker-2021-promoting,steigerwald-etal-2022-overcoming}, and adopting plain language has been shown to improve the readability of translated health content \citep{Rossetti2019}. In MT research, a wealth of studies have considered pre-processing strategies to rewrite inputs, particularly for statistical MT \citep{XiaMcCord2004,callison-burch-etal-2006-improved,stajner-popovic-2016-text}.


The growing use of Large Language Models (LLMs) for translation leads us to revisit the impact of rewriting inputs on MT. On the one hand, rewriting inputs for LLM translation aligns with the re-framing of MT as a multi-step process \citep{Briakou}. LLMs have shown promise in rewriting MT outputs \citep{ki2024guiding, zeng2024improving, xu2024llmrefine}, and can rewrite text according to various style specifications \citep{raheja-etal-2023-coedit, hallinan2023steer, shu2023rewritelm, dipper}. On the other hand, current models might already be robust to input variability, since they are trained on vast amounts of heterogeneous data \citep{touvron2023llama}, fine-tuned on diverse tasks \citep{raffel-etal-2020-exploring,alves2024tower} and operate at a much higher quality level compared to the statistical MT systems used in previous pre-processing studies.

How should inputs be rewritten for MT? The assumption that well-written texts are easier to translate drives recommendations for MT literacy, as well as the use of paraphrasing \citep{callison-burch-etal-2006-improved, mirkin-etal-2009-source, marton-etal-2009-improved, aziz-etal-2010-learning} and simplification  \citep{stajner-popovic-2016-text,stajner-popovic-2019-automated}. However, can we more directly rewrite inputs so that they are easier to translate? Generic translatability has been defined as “a measurement of the time and effort
it takes to translate a text” \citep{kumhyr-etal-1994-internationalization}. \citet{uchimoto-etal-2005-automatic} introduced a metric to quantify MT translatability based on back-translation of MT hypotheses in the source language. Given recent progress in quality estimation \citep{fernandes-etal-2023-devil, naskar-etal-2023-quality, tomani2024qualityaware}, we propose instead to use reference-free quality estimation scores as a measure of translatability.

We thus ask the following research questions:
\begin{enumerate}[label=(\arabic*),topsep=0pt,itemsep=-1ex,partopsep=-1ex,parsep=1ex]
    \item Can we improve MT quality from LLMs by rewriting inputs for style?
    \item Do quality estimation metrics provide useful translatability signals for input rewriting?
\end{enumerate}

We conduct an empirical study with 3 open-weight LLMs for a total of 21 input rewriting methods with varying levels of MT-awareness on translation from English into German, Russian and Chinese, and we further evaluate the generalizability of our best performing approach on translation from English into Czech, Hebrew and Japanese (\S \ref{sec:newlanguages}). 
Our results show that simple \textbf{MT-Agnostic rewrites} obtained by prompting LLMs to simplify, paraphrase, or change the style of the input, improve translatability, and that simplification most reliably improves translation quality. Interestingly, these MT-agnostic rewrites are more effective than \textbf{Task-Aware rewrites}, where LLMs are prompted to rewrite inputs for the purpose of MT (\S \ref{simplification best}). Finally, using quality estimation signals to assess \textbf{translatability} at the segment level and select when to use rewrites further improves MT quality, outperforming more expensive fine-tuning strategies (\S \ref{input selection}). Human evaluation further confirms that simplified rewrites and their MT largely preserve the original meaning of the source and MT (\S \ref{human evaluation}).




%

\definecolor{light blue}{RGB}{215, 242, 252}
\definecolor{light purple}{RGB}{247, 215, 252}
\definecolor{light orange}{rgb}{0.9961, 0.875, 0.7188}

\section{Input Rewriting Methods}

\label{3 method}
Within the process of source rewriting, the goal of a rewrite model is to rewrite the original source sentence $s$ into another form that is easier to translate while preserving its intended meaning. For \textbf{MT-Agnostic} rewriting methods (\S \ref{3.1 mt-agnostic}), which lacks translation-related knowledge, the rewrite model $\mathcal{M}_{\theta}$ can rewrite $s$ into $s'$:
\begin{equation}
    s' = \mathcal{M}_{\theta}(s)
\end{equation}

On the contrary, both \textbf{Task-Aware} (\S \ref{3.2 task-aware}) and \textbf{Translatability-Aware} (\S \ref{3.3 translatability-aware}) rewriting methods incorporate some translation signal. For Task-Aware, $\mathcal{M}_{\theta}$ rewrites $s$ with the information of the end-task (MT):

\begin{equation}
    s' = \mathcal{M}_{\theta}(s, \text{MT task})
\end{equation}

For Translatability-Aware method, it rewrites with the knowledge of segment level quality estimation scores between source and the output of a specific MT system MT($t$):

\begin{equation}
    s' = \mathcal{M}_{\theta}(s, \text{\textsc{xCOMET}}(s,\text{MT}(t)))
\end{equation}
Figure~\ref{fig:main_figure} shows the overview of our proposed rewriting pipeline. To find the most effective $\mathcal{M}_{\theta}$, we test a total of 21 input rewriting methods.

\subsection{\fcolorbox{white}{light blue}{\raisebox{-0.2em}{\includegraphics[height=1em]{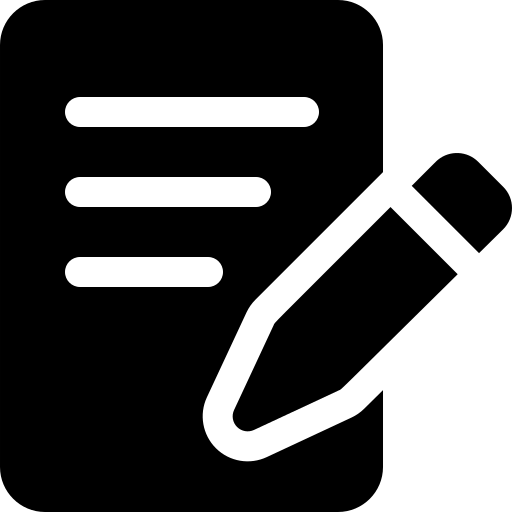}} MT-Agnostic} Rewriting}
\label{3.1 mt-agnostic}
MT-agnostic rewriting methods reflect various a priori assumptions on what makes text easier to translate. They do not take as input any signal of translatability or knowledge about the end-task. We consider three prompting variants here, all inspired by prior works on source rewriting \citep{mirkin-etal-2009-source, mirkin-etal-2013-sort, stajner-popovic-2016-text}.

\paragraph{Simplification.}
Simplification includes replacing complex words with simpler ones, rephrasing complex syntactic structures, and shortening sentences \citep{article, Feng2008}. Prior works show that simplified inputs are more conducive to MT, and particularly improve the fluency of MT outputs \citep{stajner-popovic-2019-automated}.

\paragraph{Paraphrase.}
Paraphrases are alternative ways of expressing the same information within one language, which can help resolve unknown or complex words \citep{callison-burch-etal-2006-improved}. Paraphrasing with LLMs might benefit MT by normalizing inputs using language patterns that are more frequent in LLM training data. Further, some LLMs, such as \textsc{Tower} \citep{alves2024tower}, are fine-tuned on both paraphrasing and MT tasks, and might thus produce paraphrases that are useful for MT.

\paragraph{Stylistic.}
We employ an off-the-shelf text editing tool \textsc{CoEdIT-XL} \citep{raheja-etal-2023-coedit} to rewrite inputs according to diverse style specifications:
\begin{itemize}[leftmargin=*, itemsep=2pt, parsep=-1pt]
 \item \textbf{Grammar}: Fix the grammar.
 \item \textbf{Coherent}: Make the text more coherent.
 \item \textbf{Understandable}: Make it easier to understand.
 \item \textbf{Formal}: Rewrite the text more formally.
\end{itemize}
These operationalize the assumption that well-formed text is easier to translate.
All prompt templates are shown in Appendix Table~\ref{tab:prompting_template}.

\subsection{\fcolorbox{white}{light purple}{\raisebox{-0.2em}{\includegraphics[height=1em]{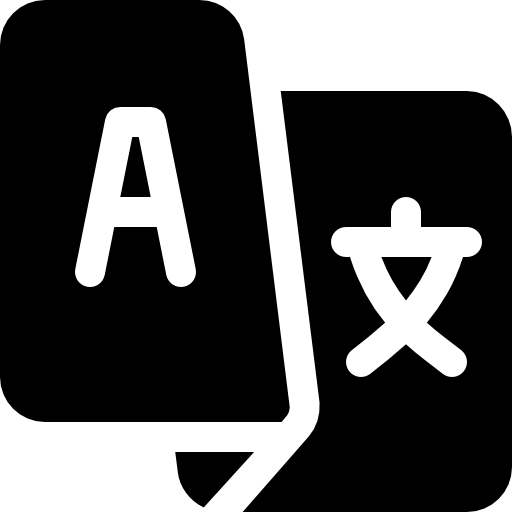}} Task-Aware} Rewriting}
\label{3.2 task-aware}

For task-aware rewriting methods, we design prompts that account for the fact that rewrites are aimed at MT. 
Prior work has shown that LLMs can post-edit errors in MT outputs \citep{ki2024guiding, zeng2024improving, treviso-etal-2024-xtower, xu2024llmrefine, briakou-etal-2024-translating}, raising the question of whether this ability can be extended to rewriting inputs to enhance translatability. Additionally, \textsc{Tower-Instruct} has been jointly trained on paraphrasing, grammatical error correction (GEC), and translation tasks, suggesting it may be well-suited for performing translatability rewrites in a zero-shot fashion. We consider two prompting strategies (Refer to Appendix Table~\ref{tab:prompting_template} for exact templates):

\paragraph{Easy Translation.} We prompt LLMs to rewrite inputs in a way that specifically facilitates translation into the target language.

\paragraph{Chain of Thought Rewrite+Translate.} We use a Chain of Thought (\citet{wei2023chainofthought}, CoT) style prompt where LLMs are prompted to handle the entire rewriting and translation process in one sequence of CoT instructions within a single model.

\subsection{\fcolorbox{white}{light orange}{\raisebox{-0.2em}{\includegraphics[height=1em]{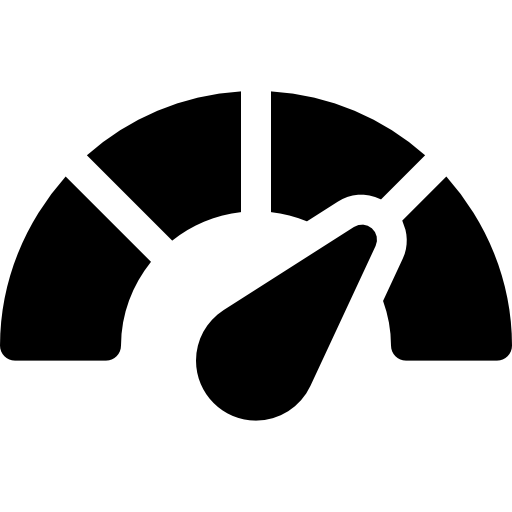}} Translatability-Aware} Rewriting}
\label{3.3 translatability-aware}
We propose to use quality estimation scores for a given input and output pair to assess the translatability of inputs at the segment level. This makes it possible to inject translatability signals at inference or training time. We introduce a lightweight inference-time selection strategy, and contrast it against a more expensive fine-tuning approach.

\paragraph{Inference-Time Selection.}
Input segments might not benefit from rewriting uniformly, since the quality of the original inputs and of their rewrites might vary. We thus propose to use translatability scores to decide whether or not to replace the original input with a rewrite  at inference time. We use the state-of-the-art \textsc{xCOMET} quality estimation tool \citep{guerreiro2023xcomet} to assess how good the translation $t'$ of a rewrite $s'$ is: \textsc{xCOMET}$(s',t')$. We compare this score with the estimated quality of the translation $t$ of the original source $s$, choosing to use the rewrite if \textsc{xCOMET}$(s',t')$ > \textsc{xCOMET}$(s,t)$, and keeping the original source otherwise. This straightforward approach allows us incorporate translatability signals at inference time, with little additional cost.


\paragraph{Supervised Fine-tuning.}
The translatability-based selection process described above for inference could also be used to gather examples of good rewrites and enable instruction fine-tuning of models to rewrite text for improved translation.
%
%
While designing an optimal approach for this task is out of scope for this work,  we wish to compare our inference-time selection strategy with a straightforward training strategy. We construct a fine-tuning dataset of positive rewrite examples $\mathcal{D}_{pos}$, as follows: for a given input $s$, we generate rewrites using all MT-agnostic methods. We add to our training set the rewrites that improve translatability as measured by \textsc{xCOMET}$(s', t')$ > \textsc{xCOMET}$(s,t)$. The base LLM is then instruction fine-tuned based to rewrite input $s$ so that it is better translated, using $s'$ as supervision. Detailed prompt templates are shown in Appendix \ref{appendix:prompt_templates}.



\section{Experimental Setup}

\definecolor{light blue}{RGB}{215, 242, 252}
\definecolor{light purple}{RGB}{247, 215, 252}
\definecolor{light orange}{rgb}{0.9961, 0.875, 0.7188}

\begin{table*}
\centering
\resizebox{\textwidth}{!}{%
\renewcommand{\arraystretch}{1.1}
\everymath{\Large}
\begin{tabular}{c l l l l l l}
    \specialrule{1.3pt}{0pt}{0pt}
    \textbf{Language} & \textbf{Type} & \textbf{Prompt/Model} & \textbf{\textsc{xCOMET}$(s,t)$} & \textbf{\textsc{xCOMET}$(s,t,r)$} & \textbf{\textsc{MetricX}$(s,t)$} & \textbf{\textsc{MetricX}$(t,r)$} \\
    \toprule
    
    \multirow{7}{*}{\Large \textbf{{\textsc{en-de}}}} & \textbf{ Original} & - & 0.893 & 0.898 & 2.038 & 1.534 \\
    
    
    & \fcolorbox{white}{light blue}{\raisebox{-0.2em}{\includegraphics[height=1em]{figures/logos/agnostic.png}} \textbf{MT-Agnostic}} & Simplification \small (\textsc{Tower}) & \textbf{0.915} & \textbf{0.907} & \textbf{1.504*} & 1.519 \\
    & & Paraphrase \small (\textsc{DIPPER}) & \textbf{0.904} & 0.838 & \textbf{1.674} & 2.757 \\
    
    
    & \fcolorbox{white}{light purple}{\raisebox{-0.2em}{\includegraphics[height=1em]{figures/logos/task.png}} \textbf{Task-Aware}} & Easy translate \small (\textsc{Tower}) & \textbf{0.901} & 0.903 & \textbf{1.759} & 2.427 \\
    & & CoT \small (\textsc{Tower}) & \textbf{0.907} & 0.897 & \textbf{1.892} & 1.578 \\

    
    & \fcolorbox{white}{light orange}{\raisebox{-0.2em}{\includegraphics[height=1em]{figures/logos/translatability.png}} \textbf{Translatability-Aware}} & Selection & \textbf{0.921*} & \textbf{0.922*} & \textbf{1.734} & \textbf{1.461*} \\
    & & Fine-tune \small (Ref) & \textbf{0.896} & 0.876 & 2.023 & 2.028 \\

    \toprule

    \multirow{7}{*}{\Large \textbf{{\textsc{en-ru}}}} & \textbf{ Original} & - & 0.872 & 0.868 & 2.535 & 2.028 \\
    
    
    & \fcolorbox{white}{light blue}{\raisebox{-0.2em}{\includegraphics[height=1em]{figures/logos/agnostic.png}} \textbf{MT-Agnostic}} & Simplification \small (\textsc{Tower}) & \textbf{0.921*} & \textbf{0.891} & \textbf{1.135} & \textbf{1.921} \\
    & & Paraphrase \small (\textsc{DIPPER}) & \textbf{0.904} & 0.821 & \textbf{1.249} & 3.476 \\
    
    
    & \fcolorbox{white}{light purple}{\raisebox{-0.2em}{\includegraphics[height=1em]{figures/logos/task.png}} \textbf{Task-Aware}} & Easy translate \small (LLaMA-3) & \textbf{0.917} & \textbf{0.881} & \textbf{0.801*} & 10.401 \\
    & & CoT \small (\textsc{Tower}) & \textbf{0.903} & 0.875 & 2.432 & 2.024 \\

    
    & \fcolorbox{white}{light orange}{\raisebox{-0.2em}{\includegraphics[height=1em]{figures/logos/translatability.png}} \textbf{Translatability-Aware}} & Selection & \textbf{0.914} & \textbf{0.899*} & \textbf{2.096} & \textbf{1.830*} \\
    & & Fine-tune \small (Ref) & \textbf{0.894} & 0.866 & \textbf{2.284} & 2.012 \\

    \toprule

    \multirow{6}{*}{\Large \textbf{{\textsc{en-zh}}}} & \textbf{ Original} & - & 0.786 & 0.794 & 3.445 & 2.282 \\
    
    
    & \fcolorbox{white}{light blue}{\raisebox{-0.2em}{\includegraphics[height=1em]{figures/logos/agnostic.png}} \textbf{MT-Agnostic}} & Simplification \small (\textsc{Tower}) & \textbf{0.821} & \textbf{0.802} & \textbf{1.521*} & 2.227 \\
    & & Paraphrase \small (\textsc{DIPPER}) & \textbf{0.813} & 0.722 & \textbf{1.583} & 4.009 \\
    
    
    & \fcolorbox{white}{light purple}{\raisebox{-0.2em}{\includegraphics[height=1em]{figures/logos/task.png}} \textbf{Task-Aware}} & Easy translate \small (LLaMA-3) & \textbf{0.793} & 0.791 & \textbf{1.618} & 7.650 \\
    & & CoT \small (\textsc{Tower}) & \textbf{0.821} & 0.771 & \textbf{3.321} & 2.432 \\

    
    & \fcolorbox{white}{light orange}{\raisebox{-0.2em}{\includegraphics[height=1em]{figures/logos/translatability.png}} \textbf{Translatability-Aware}} & Selection & \textbf{0.823*} & \textbf{0.819*} & \textbf{3.149} & \textbf{2.206*} \\    

    \specialrule{1.3pt}{0pt}{0pt}
    \end{tabular}
}
\caption{Results using different rewriting methods. Statistically significant average improvements ($p$-value $< 0.05$) are \textbf{bold}. Best scores for each metric is \textbf{bold} with \textbf{*}. \textsc{xCOMET}$(s,t)$: translatability (↑); \textsc{xCOMET}$(s,t,r)$: overall translation quality (↑); \textsc{MetricX}$(s,t)$: quality estimation (↓); \textsc{MetricX}$(t,r)$: reference-based metric (↓). We substitute $s$ and $t$ to $s'$ and $t'$ when computing scores for rewrites. For each rewriting type, we show the best and worst of each methods based on \textsc{xCOMET}$(s,t,r)$. We abbreviate \textsc{Tower-Instruct} as \textsc{Tower} and \textsc{DIPPER (L80/O60)} as \textsc{DIPPER} due to space constraints. Full results are in Appendix \ref{appendix:detailed results}.}
\label{tab:main_results}
\end{table*}

\subsection{Model \& Data}

\paragraph{MT System.} We use \textsc{Tower-Instruct} 7B as our MT system for all our experiments since it is specifically trained for translation-related tasks and has demonstrated superior MT performance compared to other LLMs \citep{alves2024tower}.

\paragraph{Rewriting Models.} For prompting experiments, we use 7B variant of three open-weight LLMs in zero-shot setting: \textsc{LLaMA-2} \citep{touvron2023llama} \---\ the base model for \textsc{Tower-Instruct}, \textsc{LLaMA-3} \citep{grattafiori2024llama3herdmodels} \---\ more recent multilingual model compared to \textsc{LLaMA-2}, and \textsc{Tower-Instruct} \citep{alves2024tower} \---\ the same LLM as used for our MT system.\footnote{The HuggingFace model names are detailed in Appendix Table \ref{tab:huggingface_api}.} For supervised fine-tuning, we draw training samples from the English-German and English-Russian subset from WMT-20, 21, and 22 General MT task datasets \citep{freitag2021experts}\footnote{We do not consider English-Chinese pair here since this language pair is not supported in the dataset.}, and provide detailed parameter settings in Appendix~\ref{appendix:parameters}.


\paragraph{Test Data.}
We use the WMT-23 General MT task\footnote{\url{https://www2.statmt.org/wmt23/translation-task.html}} from the \textsc{TowerEval} dataset\footnote{\url{https://huggingface.co/datasets/Unbabel/TowerEval-Data-v0.1}} to guarantee that it was held out from the various training stages. We focus on translation from English into German (\textsc{En-De}), Russian (\textsc{En-Ru}) and Chinese (\textsc{En-Zh}) for an extensive empirical comparison, and then test whether the most promising approaches generalize to translation from English into Czech (\textsc{En-Cs}), Hebrew (\textsc{En-He}) and Japanese (\textsc{En-Ja}). 
See Appendix Table~\ref{tab:dataset_details} for data statistics.



\subsection{Evaluation Metrics}
We use \textsc{xCOMET} \citep{guerreiro2023xcomet} and \textsc{MetricX} \citep{juraska-etal-2023-metricx} to evaluate different aspects of rewrite quality. Specifically, we use \textsc{xCOMET-XL}\footnote{\url{https://huggingface.co/Unbabel/XCOMET-XL}} and \textsc{MetricX-23-XL}.\footnote{\url{https://huggingface.co/google/metricx-23-xl-v2p0}} Higher scores indicate better performance for \textsc{xCOMET}, while lower scores are better with \textsc{MetricX}.

\paragraph{Translatability.}
We quantify translatability with the quality estimation score for a specific input--output pair (\textsc{xCOMET}$(s', t')$ or \textsc{MetricX-QE}$(s',t')$). A rewrite $s'$ of the original input $s$ is considered easier to translate if \textsc{xCOMET}$(s', t')$ is higher than \textsc{xCOMET}$(s, t)$.

\paragraph{Meaning Preservation.} We do not want rewrites that are easier to translate at the expense of changing the original meaning. Our meaning preservation metric evaluates how well the rewrite maintains the intended meaning of the translation as represented by the reference \citep{Graham2015CanMT}. We use a reference-based metric as opposed to using the semantic similarity between $s$ and $s'$ because it abstracts the meaning away from the specific formulation of $s$, reducing overfitting. We compute \textsc{xCOMET} scores between the rewrites and reference translations (\textsc{xCOMET}$(s',r)$). The desired behavior is to minimize the deterioration in \textsc{xCOMET}$(s',r)$ compared to \textsc{xCOMET}$(s,r)$.

\paragraph{Translation Quality.} We additionally report the combined evaluation metric, \textsc{xCOMET}$(s',t',r)$ to take into account of the trade-off between the two above metrics, and \textsc{MetricX}$(t',r)$ which also assesses translation quality of the rewrite but is not informed by the updated source $s'$.

\definecolor{light blue}{RGB}{215, 242, 252}
\definecolor{light purple}{RGB}{247, 215, 252}
\definecolor{light orange}{rgb}{0.9961, 0.875, 0.7188}

\section{Results}
\label{4 results}

We first extensively compare rewrite strategies focusing on the overall translation quality achieved by MT-Agnostic rewrites (\S \ref{simplification best}) and Translatability-Aware rewrites (\S \ref{input selection}). To understand how rewrites change translations, we then analyze the trade-offs between translatability and meaning preservation (\S \ref{pareto optimality}). Finally, we test whether the best-performing methods identified so far generalize to new language pairs (\S \ref{sec:newlanguages}).

\subsection{Simplifying Inputs Works Best}
\label{simplification best}
We first compare the \hl{MT Agnostic} rewriting methods: simplification, paraphrasing, and stylistic edits. Due to space limits, we show the best and worst performing variations for each input rewriting method based on the overall translation quality metric \textsc{xCOMET}$(s,t,r)$ for each language pair in Table~\ref{tab:main_results}. Full results are available in Appendix \ref{appendix:detailed results}.

Results show that all rewriting strategies improve translatability, but only \textbf{simplification} also improves the overall translation quality. Even the lowest performing rewrites reach higher translatability than the original baseline. Each method surpasses the baseline by up to 0.056 and 0.027 \textsc{xCOMET}$(s,t)$ average scores for \textsc{En-De}, up to 0.058 and 0.036 average scores for \textsc{En-Ru}, and up to 0.054 and 0.028 average scores for \textsc{En-Zh} pair. Trends are consistent with \textsc{MetricX}$(s,t)$. However, making inputs easier to translate often degrades quality when comparing against references $r$. Simplification with \textsc{Tower-Instruct} distinguishes itself by improving translation quality based on \textsc{xCOMET}$(s,t,r)$ scores and maintaining it according to the \textsc{MetricX}$(t,r)$ scores \---\ a harder metric to improve since the reference might be biased toward the original wording of the source.


Among the three LLMs used for simplification, \textsc{Tower-Instruct} achieves the best translation quality, while \textsc{LLaMA-3} excels in translatability at the expense of meaning preservation. Interestingly, there is no benefit to using a separate LLM, even one fine-tuned specifically on paraphrasing or style edits such as \textsc{DIPPER} or \textsc{CoEdIT}. Overall, the best performing method for MT-agnostic rewrites is simplification with \textsc{Tower-Instruct}, the same model we use as our MT system. We attribute this to \textsc{Tower-Instruct} being instruction fine-tuned on translation related tasks (but not simplification) and having more domain knowledge of the WMT dataset used in our evaluation.\footnote{\url{https://huggingface.co/datasets/Unbabel/TowerBlocks-v0.1}}


As shown in Table~\ref{tab:main_results}, simplifying with \textsc{Tower-Instruct} still holds the top spot when compared to \hlpurple{Task-Aware} rewriting methods, as indicated by higher \textsc{xCOMET}$(s,t,r)$ scores. This suggests that injecting knowledge about the end-task (MT) to LLMs is less effective than simplifying inputs to improve translation quality.

Overall, these results confirm the intuition that simpler text is easier to translate, but establish that rewrites are not uniformly helpful for translation quality, motivating the need for more selective input rewriting strategies.


\begin{figure*}
    \centering
    \includegraphics[width=\linewidth]{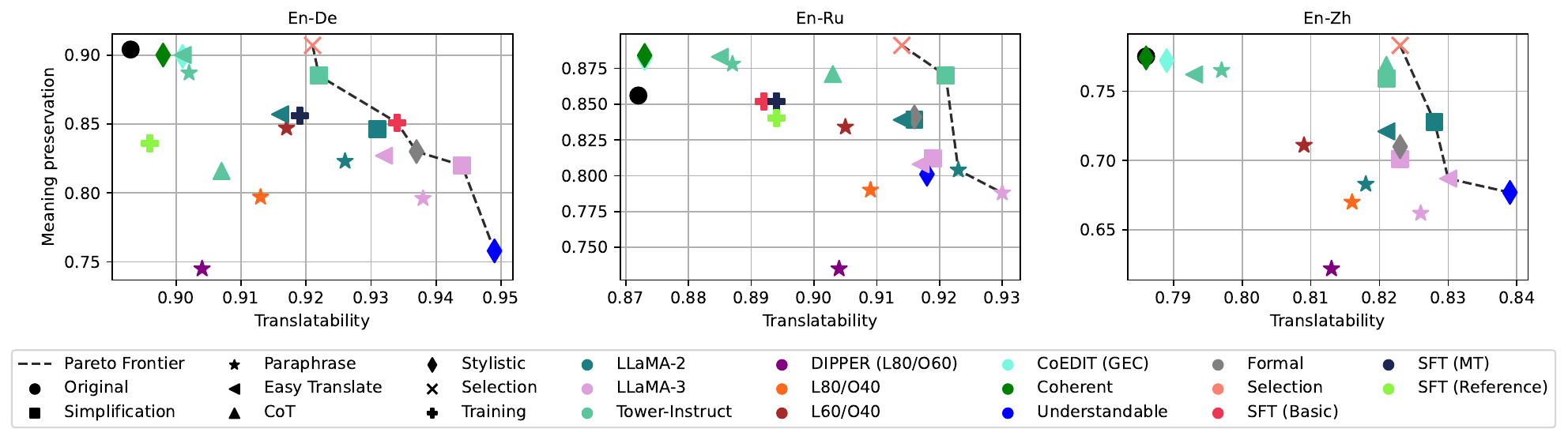}
    \caption{Pareto frontier per language pair. For each subplot, the $x$-axis is the translatability and $y$-axis is the meaning preservation scores. Pareto frontier (\textbf{dashed} line) visualizes the optimal solutions that take into account the trade-off between the two metrics. Each shape represents different rewriting methods and each color represent specific prompt or model variation.}
    \label{fig:pareto_frontier}
\end{figure*}

\subsection{Selection via Translatability Improves MT}
\label{input selection}

We evaluate the impact of inference-time selection based on \hlorange{translatability} scores (\textit{Selection} in Table~\ref{tab:main_results}), and compare it further with the more expensive supervised fine-tuning strategy (\textit{Fine-tune}). 

All language pairs consistently benefit from selection. Translation quality improves significantly, with average \textsc{xCOMET}$(s,t,r)$ gains of 0.024 for \textsc{En-De}, 0.031 for \textsc{En-Ru}, and 0.025 for \textsc{En-Zh}, marking the best performance among all variants. \textsc{MetricX}$(t,r)$ scores confirm this trend, showing average improvements of 0.073 for \textsc{En-De}, 0.198 for \textsc{En-Ru}, and 0.076 for \textsc{En-Zh}. At the segment level, rewrites are preferred to original inputs in 1197/1557 cases for \textsc{En-De}, 1610/2074 cases for \textsc{En-Ru}, and 2163/3074 cases for \textsc{En-Zh}. Fine-tuning shows smaller gains compared to MT-Agnostic or Task-Aware methods, both in terms of translatability and translation quality, despite being more resource-intensive.

In summary, the results suggest that inference-time selection of inputs based on translatability scores is a promising strategy, outperforming MT-agnostic rewrites and rewrites obtained via a more expensive fine-tuning process.

\subsection{Input Rewriting Trades Off Translatability and Meaning Preservation}
\label{pareto optimality}

We observe a moderate negative correlation between translatability and meaning preservation scores, with Pearson coefficients of -0.48, -0.66, and -0.52 for \textsc{En-De}, \textsc{En-Ru}, and \textsc{En-Zh}, respectively. This trade-off between the two metrics poses a Pareto optimization challenge: when a rewrite is easier to translate, it often results in lower meaning preservation. Therefore, we aim to find Pareto optimal solutions, which balance these trade-offs on a Pareto frontier \citep{huang-etal-2023-towards}.\footnote{In Pareto optimization, Pareto optimal solutions are those where no single solution outperforms another in all tasks \citep{pareto}. The set of Pareto optimal solutions forms the Pareto frontier.}

In Figure~\ref{fig:pareto_frontier}, we visualize our two objectives, translatability and meaning preservation, on each axis and identify the Pareto frontier. The results are consistent with the overall translation quality metric, \textsc{xCOMET}$(s,t,r)$, where the scores for rewriting methods on the Pareto frontier are consistently the same as or on par with the original baseline. This also aligns with our earlier findings from comparing MT-Agnostic and Task-Aware rewrites (\S \ref{simplification best}), where simplification with \textsc{Tower-Instruct} lies on the Pareto frontier for \textsc{En-De} and \textsc{En-Ru}. Even for \textsc{En-Zh}, although this does not lie on the frontier, it has a higher \textsc{xCOMET}$(s,t,r)$ score (0.802) than the original baseline (0.794). Furthermore, the best rewriting method according to \textsc{xCOMET}$(s,t,r)$, translatability-based selection (\S \ref{input selection}), always lies on the Pareto frontier across all language pairs.

\begin{table}[!htp]
\centering
\resizebox{\linewidth}{!}{%
    \begin{tabular}{l l l l l l l}
    \specialrule{1.3pt}{0pt}{0pt}
    \textbf{Language} & \textbf{Type} & \textbf{\textsc{x}$(s,t)$} & \textbf{\textsc{x}$(s,t,r)$} & \textbf{\textsc{M}$(s,t)$} & \textbf{\textsc{M}$(t,r)$} \\
    \toprule
        
    \multirow{3}{*}{\textbf{\large{\textsc{en-cs}}}} & Original & 0.646 & 0.655 & 5.376 & 4.493 \\
    & Simplification & 0.691 & 0.675 & 4.684 & 4.333\\
    & Selection & \textbf{0.736} & \textbf{0.718} & \textbf{4.152} & \textbf{3.663}\\ \midrule

    \multirow{3}{*}{\textbf{\large{\textsc{en-he}}}} & Original & 0.327 & 0.320 & 16.66 & 15.48 \\
    & Simplification & 0.351 & 0.332 & 15.97 & 15.43 \\
    & Selection & \textbf{0.389} & \textbf{0.363} & \textbf{15.39} & \textbf{14.51} \\ \midrule

    \multirow{3}{*}{\textbf{\large{\textsc{en-ja}}}} & Original & 0.746 & 0.718 & 3.514 & 2.688 \\
    & Simplification & 0.789 & 0.738 & 2.957 & 2.508 \\
    & Selection & \textbf{0.826} & \textbf{0.769} & \textbf{2.781} & \textbf{2.273} \\



    \specialrule{1.3pt}{0pt}{0pt}
    \end{tabular}
}
\caption{Results of simplification and translatability-based selection for held-out test sets. We abbreviate \textsc{xCOMET} to \textbf{\textsc{x}} and \textsc{MetricX} to \textbf{\textsc{M}} due to space constraints. Best scores for each metric is \textbf{bold}.} 
\label{tab:heldout}
\end{table}

\subsection{Best Input Rewriting Strategy Improves MT on Held-out Test sets}
\label{sec:newlanguages}

We evaluate whether the top methods that have emerged from the controlled empirical comparison conducted so far generalize to further test settings. As shown in Table~\ref{tab:heldout}, we test both simplification with \textsc{Tower-Instruct} (\textit{Simplification}) and translatability-based input selection (\textit{Selection}) on new test sets from the  WMT-23 General MT task, English-Czech (\textsc{En-Cs}), English-Hebrew (\textsc{En-He}), and English-Japanese (\textsc{En-Ja}) to assess generalization to lower-resource target languages.

Both simplification and translatability-based selection lead to progressive improvements in translation quality, as measured by \textsc{xCOMET}$(s,t,r)$. Notably, the selection strategy tends to excel in language pairs with lower-resource target languages, showing translation quality gains of 0.064, 0.043, 0.051 scores for \textsc{En-Cs}, \textsc{En-He}, \textsc{En-Ja}, respectively, compared to increases of 0.017, 0.031, and 0.025 for \textsc{En-De}, \textsc{En-Ru} and \textsc{En-Zh}. At the segment level, rewrites are also more preferred over original inputs, selected in 1395/2074 cases for \textsc{En-Cs}, 1309/2074 for \textsc{En-He}, and 1411/2074 for \textsc{En-Ja}. \textsc{MetricX} trends are consistent.

In sum, our findings generalize well to held-out test sets, further validating the effectiveness of the translatability-based selection strategy. This approach offers a practical and scalable solution for input rewriting across a broader range of domains and language pairs, though there are many other dimensions that remain unexplored. We have conducted initial experiments with additional LLMs and source languages, shown in Appendix \ref{appendix:more_llms} and \ref{appendix:more_lang_pairs}, which confirms our previous findings that simplification rewriting enhances translation quality. We leave a more comprehensive exploration of this direction for future work.

\section{Analysis}
\label{5 analysis}

\subsection{Simplifying Inputs Improves MT Readability}
\label{readability}

Simplification as an input rewriting strategy can balance translatability and meaning preservation, leading to overall improvements in translation quality. We also examine whether this enhances the readability of both inputs and, subsequently, translation outputs. In Table~\ref{tab:readability}, we present the Flesch Reading Ease score\footnote{\url{https://en.wikipedia.org/wiki/Flesch-Kincaid_readability_tests}} and Gunning Fog index\footnote{\url{https://en.wikipedia.org/wiki/Gunning_fog_index}} to measure input readability, and the Vienna formula (WSTF) \citep{zowalla2023readability} and the Russian version of Flesch Readability test \citep{inbook} to assess output readability for \textsc{En-De} and \textsc{En-Ru}, respectively.

As expected, input readability improves across all simplification methods, whether used in MT-Agnostic (\textsc{LLaMA-2}, \textsc{LLaMA-3}, and \textsc{Tower-Instruct} in Table~\ref{tab:readability}) or Translatability-Aware (Selection in Table~\ref{tab:readability}) manner. Interestingly, simplification not only leads to more readable input but also more readable outputs, with gains of up to 0.22 WSTF scores for \textsc{En-De} and 0.95 Flesch scores for \textsc{En-Ru}. We provide several qualitative examples in Appendix Tables \ref{tab:readability_ende} to \ref{tab:readability_enzh} that illustrate how simplification rewrites can lead to varying degrees of readability improvements in both inputs and translation outputs.

\begin{table}[!htp]
\centering
\resizebox{\linewidth}{!}{%
    \begin{tabular}{l l l l l l}
    \specialrule{1.3pt}{0pt}{0pt}
    \textbf{Language} & \textbf{Prompt/Model} & \textbf{Flesch} & \textbf{GFI} & \textbf{WSTF} & \textbf{Flesch-Ru} \\
    \toprule

    \multirow{5}{*}{\large \textbf{\Large{\textsc{en-de}}}} & Original & 60.79 & 10.56 & 1.35 & - \\
    & \textsc{LLaMA-2} & 66.69 & 9.25 & 1.15 & - \\
    & \textsc{LLaMA-3} & 64.00 & 9.98 & 1.24 & - \\
    & \textsc{Tower-Instruct} & \textbf{68.17} & \textbf{8.99} & \textbf{1.13} & - \\
    & Selection & 63.27 & 10.09 & 1.26 & - \\ \midrule

    \multirow{5}{*}{\large \textbf{\Large{\textsc{en-ru}}}} & Original & 69.93 & 9.91 & - & 65.67 \\
    & \textsc{LLaMA-2} & \textbf{74.73} & 8.37 & - & \textbf{66.62} \\
    & \textsc{LLaMA-3} & 72.88 & 9.20 & - & 66.36 \\
    & \textsc{Tower-Instruct} & 74.14 & \textbf{8.19} & - & 65.40 \\
    & Selection & 72.24 & 9.37 & - & 65.89 \\ \midrule

    \multirow{5}{*}{\large \textbf{\Large{\textsc{en-zh}}}} & Original & 66.51 & 10.08 & - & - \\
    & \textsc{LLaMA-2} & 71.64 & 8.74 & - & - \\
    & \textsc{LLaMA-3} & 69.32 & 9.48 & - & - \\
    & \textsc{Tower-Instruct} & \textbf{72.22} & \textbf{8.42} & - & - \\
    & Selection & 68.41 & 9.68 & - & - \\

    \specialrule{1.3pt}{0pt}{0pt}
    \end{tabular}
}
\caption{Input and output readability scores for simplification rewriting method. \textbf{Flesch}: Flesch Reading Ease score (↑); \textbf{GFI}: Gunning Fog Index (↓); \textbf{WSTF}: Vienna formula (↓); \textbf{Flesch-Ru}: Russian version of Flesch (↑).
}
\label{tab:readability}
\end{table}
\begin{table}[!htp]
\centering
\resizebox{\linewidth}{!}{%
    \begin{tabular}{l l l l l l l}
    \specialrule{1.3pt}{0pt}{0pt}
    \textbf{Language} & \textbf{Type} & \textbf{\textsc{x}$(s,t)$} & \textbf{\textsc{x}$(s,t,r)$} & \textbf{\textsc{M}$(s,t)$} & \textbf{\textsc{M}$(t,r)$}\\
    \toprule
        
    \multirow{5}{*}{\textbf{\large{\textsc{en-de}}}} & Original & 0.893 & 0.898 & 2.038 & 1.534 \\
    & \textbf{I} & \textbf{0.922} & \textbf{0.907} & \textbf{1.504} & 1.519 \\
    & \textbf{Owo} & 0.863 & 0.879 & 2.941 & 2.200 \\ 
    & \textbf{Ow} & 0.879 & 0.894 & 2.515 & 1.858 \\
    & \textbf{I+O} & 0.915 & \textbf{0.907} & 1.751 & \textbf{1.502} \\ \midrule

    \multirow{5}{*}{\textbf{\large{\textsc{en-ru}}}} & Original & 0.861 & 0.854 & 2.535 & 2.028 \\
    & \textbf{I} & \textbf{0.921} & 0.891 & \textbf{1.135} & \textbf{1.921} \\
    & \textbf{Owo} & 0.868 & 0.864 & 2.815 & 2.384 \\ 
    & \textbf{Ow} & 0.872 & 0.869 & 2.674 & 2.259 \\
    & \textbf{I+O} & 0.917 & \textbf{0.892} & 1.632 & 2.045 \\ \midrule

    \multirow{5}{*}{\textbf{\large{\textsc{en-zh}}}} & Original & 0.786 & 0.794 & 3.445 & \textbf{2.282} \\
    & \textbf{I} & \textbf{0.821} & 0.802 & \textbf{1.521} & 2.327 \\
    & \textbf{Owo} & 0.713 & 0.751 & 5.585 & 4.262 \\ 
    & \textbf{Ow} & 0.746 & 0.780 & 4.363 & 2.676 \\
    & \textbf{I+O} & 0.818 & \textbf{0.804} & 3.335 & 2.323 \\

    \specialrule{1.3pt}{0pt}{0pt}
    \end{tabular}
}
\caption{Results for input rewriting (\textbf{I}), post-editing output without source signal (\textbf{Owo}), with source signal (\textbf{Ow}), and the combination of both strategies (\textbf{I+O}). Best scores for each metric is \textbf{bold}. We use the same abbreviations for metrics as in Table \ref{tab:heldout}.} 
\label{tab:complementary}
\end{table}

\begin{figure*}
    \centering
    \includegraphics[width=\linewidth]{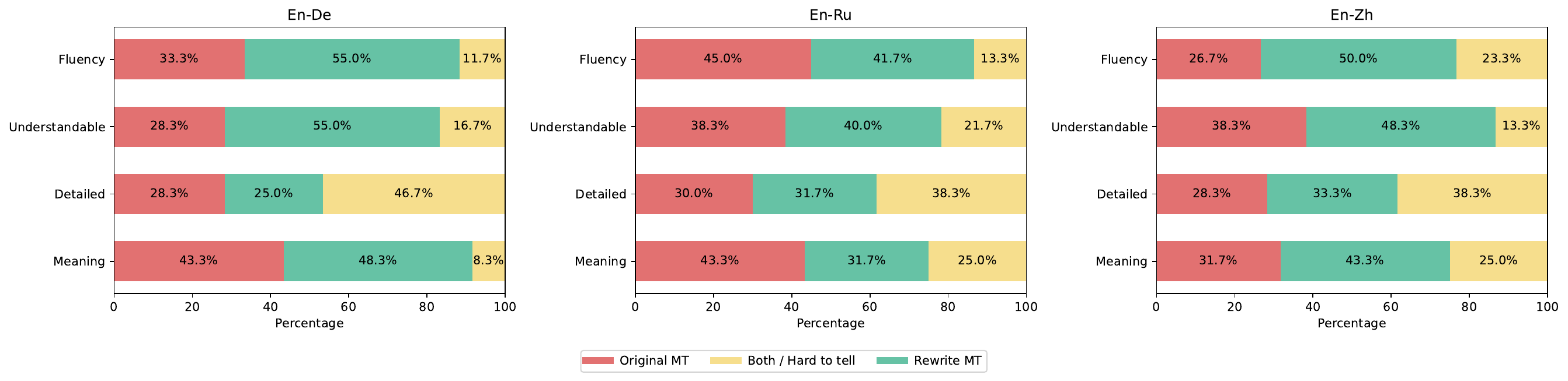}
    \caption{Win rates for human evaluation comparing Original MT vs. Rewrite MT across three language pairs (\textsc{En-De}, \textsc{En-Ru}, \textsc{En-Zh}) and four evaluation criteria: Fluency, Understandability, Level of detail (Detailed), and Meaning preservation relative to the reference translation.}
    \label{fig:human_results}
\end{figure*}

\subsection{Input Rewriting outperforms Post-Editing}
\label{res:post-editing}

The symmetric task to input rewriting is post-editing, which focuses on improving and correcting errors in translation outputs. Can post-editing alone achieve the same improvements, or are both strategies \textit{complementary}? To explore this, we compare input rewriting to post-editing by prompting \textsc{Tower-Instruct}\footnote{We focus on \textsc{Tower-Instruct} as it is a multilingual LLM capable of rewriting in non-English target languages.} to simplify either inputs or outputs. As shown in Table~\ref{tab:complementary}, rewriting inputs (\textbf{I}) offers a notable advantage over post-editing outputs (\textbf{Owo}), even when post-editing is guided by the input sentence (\textbf{Ow}). Combining input rewriting and post-editing (\textbf{I+O}) yields the highest translation quality, though the difference compared to input rewriting alone is not statistically significant. This confirms that rewriting text for better translatability before translation plays a more decisive role than post-editing the output.\footnote{We compare time and computational efficiency for input rewriting and output post-editing in Appendix \ref{appendix:inference_cost}.}

\subsection{Human Evaluation}
\label{human evaluation}

\paragraph{Original MT vs. Rewrite MT.}
We conduct a manual evaluation to determine whether bilingual human annotators rate translations generated using our winning rewrite method (simplification with \textsc{Tower-Instruct}) as superior to the original translations. For each language pairs (\textsc{En-De}, \textsc{En-Ru}, \textsc{En-Zh}), we randomly select 20 pairs of instances, resulting in a total of 180 annotations from three annotators per pair. Inter-annotator agreement, measured by Fleiss' Kappa\footnote{\url{https://en.wikipedia.org/wiki/Fleiss_kappa}}, is moderate, with values of 0.43, 0.39, and 0.51 for \textsc{En-De}, \textsc{En-Ru}, and \textsc{En-Zh}, respectively. For each instance, annotators are first provided with two translations and asked to evaluate on three axes: \textbf{1)} Fluency, \textbf{2)} Understandability, and \textbf{3)} Level of detail. Subsequently, we provide the reference translation, and annotators are asked to assess \textbf{4)} Meaning preservation. Annotators are also given the option to provide free form comments. Further details on the annotation set-up are available in Appendix \ref{appendix:mt_details}.

As illustrated in Figure~\ref{fig:human_results}, the human evaluation results confirm that translations from simplified inputs are rated as more fluent, understandable, and better at preserving the meaning of the reference translation. While this improvement is clear for the \textsc{En-De} and \textsc{En-Zh} pair, for \textsc{En-Ru} pair, annotators rate original MT as more fluent and more faithful to the original meaning.\footnote{Note that the Fleiss's Kappa scores indicate that there is more disagreement between annotators for \textsc{En-Ru} pair.} Some \textsc{En-Ru} annotators who preferred the original MT noted that it often retained a more accurate sense of the words in the reference. In contrast, those who favored the simplified rewrite MT highlighted that translations are more contextually appropriate, easier to read, and more comprehensible than the original MT.

\paragraph{Original vs. Rewrite.}
Our automatic meaning preservation metric evaluates the extent to which the original meaning is retained in the rewrite by comparing the rewritten source to the reference translation, rather than to the original source \citep{Graham2015CanMT}. Comparing to the original source is in the same language, but introduces a bias toward the original wording. On the other hand, comparing to the reference involves a cross-lingual comparison and is affected by unstable quality of references \citep{kocmi-etal-2022-findings}, but is less biased toward the original wording of the source.

To complement our automatic metric, we conduct a manual evaluation to assess how well the rewrites from simplification with \textsc{Tower-Instruct} preserve the meaning of the original source. We randomly sample 30 pairs of instances and collect three annotations per pair, totaling 90 annotations. Annotators are presented with both the original and rewritten sources and asked to evaluate how well the rewrite captures the meaning of the original source using a 4-point Likert scale (1: Does not capture meaning, 2: Partially, 3: Mostly, 4: Fully). Inter-annotator agreement by Fleiss' Kappa is 0.45. Of the 90 annotations, 55 were rated as 4, 27 as 3, 7 as 2, and 1 as 1, resulting an average score of 3.51. These results indicate that simplified rewrites generated by \textsc{Tower-Instruct}, although compared against the original source, still largely preserve the original meaning. Further details are provided in Appendix \ref{appendix:details}.
\section{Related Work}

\paragraph{Rewriting with LLMs.} 
Recent advances in LLMs have demonstrated impressive zero-shot capabilities in rewriting textual input based on user requirements \citep{shu2023rewritelm}. Most LLM-assisted rewriting tasks focus on query rewriting \citep{efthimiadis1996query}, which aims to reformulate text-based queries to enhance their representativeness and improve recall with retrieval-augmented LLMs \citep{mao-etal-2023-search, Zhu_2024}. Rewriting methods include prompting LLMs both as rewriters and rewrite editors \citep{ye-etal-2023-enhancing, kunilovskaya-etal-2024-mitigating}, and training LLMs as rewriters using feedback alignment learning \citep{ma-etal-2023-query, mao2024rafe}. Another line of work focuses on style transfer, where the goal is to rewrite textual input into a specified style \citep{wordcraft, hallinan2023steer}. Our research aligns with efforts to rewrite texts with LLM assistance; however, unlike these works, we focus on rewriting source inputs to enhance MT quality.

\paragraph{Quality Estimation Metrics.}
The discrepancy between lexical-based metrics (e.g., \textsc{BLEU} \citep{papineni-etal-2002-bleu}, \textsc{chrF} \citep{popovic-2015-chrf}) and human judgments \citep{ma-etal-2019-results} has led to research in \textit{neural} metrics. Particularly, quality estimation (QE) metrics, which compute a quality score for the translation conditioned only on the source sentence, have demonstrated benefits in improving MT quality. QE metrics are used for various purposes, including filtering out low-quality translations during training \citep{tomani2024qualityaware}, applying to post-editing workflows \citep{bechara2021role}, and providing feedback to users of MT systems \citep{mehandru2023physician}. In our experiments, we use \textsc{xCOMET} as our main evaluation metric, as it shows the best correlation with human judgments \citep{agrawal2024automatic}. We primarily use \textsc{xCOMET} as a QE metric to compute translatability, further providing this information as knowledge to LLMs to improve MT quality.

\paragraph{Rewriting MT Outputs.} 
The symmetric task of post-editing MT outputs has received significantly more attention than rewriting MT inputs. Most recent work relies on LLMs to automatically detect and correct errors in MT outputs using their internal knowledge \citep{raunak-etal-2023-leveraging, zeng2024improving, chen2024iterative}, with the help of external feedback \citep{ki2024guiding, xu2024llmrefine} or through fine-tuning \citep{treviso2024xtowermultilingualllmexplaining}. In contrast, the task of rewriting MT inputs to make them more suitable for translation has been relatively underexplored with LLMs. While there have been some efforts in query rewriting and style transfer to improve retrieval \citep{mao-etal-2023-search, Zhu_2024} and stylistic coherence \citep{ye-etal-2023-enhancing, hallinan2023steer}, the specific application of LLMs to rewrite inputs for the purpose of enhancing MT quality is still emerging. Our research addresses this gap by focusing on the potential of LLM-assisted input rewriting to improve the translatability and quality of the resulting translations.

\section{Conclusion}

In this work, we studied the effectiveness of automatic input rewriting with LLMs in improving the quality of machine translation outputs. We explored a range of rewriting strategies with varying levels of MT-awareness: \textbf{1)} MT-Agnostic, \textbf{2)} Task-Aware (knowledge of the end-task), and \textbf{3)} Translatability-Aware rewrites (knowledge of translatability as measured with QE tools).

Our findings show that simpler texts are more translatable. However, MT-Agnostic rewrites do not uniformly help translation quality (\S \ref{simplification best}), which motivates us to explore more selective strategies. Selecting inputs based on translatability scores during inference time further boosts translation quality (\S \ref{input selection}), addressing the Pareto optimization challenge by striking a balance between translatability and meaning preservation (\S \ref{pareto optimality}). Analysis shows that simplifying inputs also results in more readable translation outputs (\S \ref{readability}), and that input rewriting complements post-editing strategies (\S \ref{res:post-editing}). Human evaluation complements our automatic metric by showing that both simplified rewrites and their corresponding MT largely preserve the original meaning of the source and MT (\S \ref{human evaluation}).

More broadly, this work suggests that LLM-assisted input rewriting is a promising direction for improving translations. The approaches introduced here represent a first step in this direction, and future work is needed to discover optimal rewriting strategies for a broader range of models. Furthermore, in line with growing research on LLM-based writing assistants \citep{Lee_2024}, these results encourage future work on designing richer interactive approaches to translation with LLMs.


\section{Limitations}


We focus our investigation on \textsc{Tower-Instruct 7B} as our MT system, as it is an open-weight model. We exclude closed and larger models such as \textsc{GPT-4}\footnote{\url{https://openai.com/index/gpt-4/}} in the current experiments.

The scope of our study is also limited to out-of-English language pairs, as rewriting with LLMs has been more extensively studied in English \citep{ma-etal-2023-query, ye-etal-2023-enhancing, shu2023rewritelm, mao2024rafe}, and using English as the source language benefits performance from its prevalence in LLM training data. One critical area of future research lies in developing rewriting tools that support a wider range of languages beyond English. 

\section*{Acknowledgments}

We thank the anonymous reviewers and the members of the \textsc{clip} lab at University of Maryland for their constructive feedback. This work was supported in part by NSF Fairness in AI Grant 2147292, by the Institute for Trustworthy AI in Law and Society (TRAILS), which is supported by the National Science Foundation under Award No. 2229885, and by the Office of the Director of National Intelligence (ODNI), Intelligence Advanced Research Projects Activity (IARPA), via the HIATUS Program contract \#2022-22072200006, by NSF grant 2147292. The views and conclusions contained herein are those of the authors and should not be interpreted as necessarily representing the official policies, either expressed or implied, of ODNI, IARPA, NSF or the U.S. Government. The U.S. Government is authorized to reproduce and distribute reprints for governmental purposes notwithstanding any copyright annotation therein.

\bibliography{custom,inputrewrite}

\appendix

\section{Model and Experiment Details}
\subsection{Prompt Templates}
\label{appendix:prompt_templates}
In Tables \ref{tab:prompting_template} and \ref{tab:training_template}, we describe the prompt templates used for prompting and fine-tuning experiments, respectively. For stylistic rewriting, we use the same prompts as those used to train the \textsc{CoEdIT-XL} model. During prompting, we provide the original source as the input, while for fine-tuning, we provide the positive rewrite along with the source.

\subsection{Training Setup}
\label{appendix:parameters}
All models are trained using one NVIDIA RTX A5000 GPU. In practice, we find that fine-tuning converges in around 3 hours. We use a 90/10 train/validation data split and adopt QLoRA \citep{dettmers2023qlora}, a quantized version of LoRA \citep{hu2021lora}, for parameter-efficient training. We train \textsc{Tower-Instruct 7B} with 8-bit quantization, a LoRA rank of 16, a scaling parameter ($\alpha$) of 32, and a dropout probability of 0.05 for layers. We train for 10 epochs. All unspecified hyperparameters are set to default values.

\subsection{Decoding Strategy}
We use greedy decoding (no sampling) when generating rewrites for prompting experiments. We fix the temperature value to 0 throughout the experiments in order to eliminate sampling variations.

\begin{table}
\centering
\resizebox{\linewidth}{!}{%
    \begin{tabular}{ll}
    \specialrule{1.3pt}{0pt}{0pt}
    \textbf{Model} & \textbf{HuggingFace Model Name} \\
    \toprule

    \textsc{LLaMA-2} & \texttt{meta-llama/Llama-2-7b-chat-hf} \\
    \textsc{LLaMA-3} & \texttt{meta-llama/Meta-Llama-3-8B-Instruct} \\
    \textsc{Tower-Instruct} & \texttt{Unbabel/TowerInstruct-7B-v0.1} \\
    \specialrule{1.3pt}{0pt}{0pt}
    \end{tabular}
}
\caption{HuggingFace model names for all tested LLMs.} 
\label{tab:huggingface_api}
\end{table}

\subsection{Dataset Details}
\label{appendix:dataset_details}
We provide detailed statistics of our training ($\mathcal{D}_{pos}$) and test dataset in Table \ref{tab:dataset_details}. For $\mathcal{D}_{pos}$, we only use rewrites where the \textsc{xCOMET}$(s', t')$ score is higher than the original \textsc{xCOMET}$(s, t)$ score. We further conduct a two-step pre-processing procedure: \textbf{1)} Remove duplicate instances and \textbf{2)} Remove lengthy instances where the upper threshold is set as Q$3 + 1.5 \times \text{IQR}$.

\section{Detailed Results}
\subsection{Full Results}
\label{appendix:detailed results}
In Tables \ref{tab:detailed_results_ende} to \ref{tab:detailed_results_enzh}, we present the detailed numerical results for all tested variations. Most rewrites yield higher \textsc{xCOMET}$(s,t)$ scores, indicating better translatability compared to the original baseline. For stylistic rewrites with \textsc{CoEdIT}, prompting to make the text easier to understand (Understandable) achieves the highest translatability score, while prompting to rewrite the text more formally (Formal) results in the highest translation quality. The Coherent prompt achieves the highest meaning preservation score but this is because most rewrites are merely copies of the original source (Appendix \ref{appendix:direct_copy}). Overall, we demonstrate that translatability-based selection method remains the most effective method, even outperforming scores from our fine-tuned LLMs.

\subsection{Impact of LLM}
\label{appendix:impact of llm}
Among the three LLMs used for prompting, \textsc{Tower-Instruct} performs the best in terms of the combined metric \textsc{xCOMET}$(s,t,r)$. Although it lags behind \textsc{LLaMA-2} and \textsc{LLaMA-3} in translatability, its meaning preservation score deteriorates the least, resulting in the highest overall score. \textsc{LLaMA-3} performs the best in terms of translatability, likely due to its more multilingual training data, with over 5\% of its pre-training dataset consisting of high-quality non-English data.\footnote{\url{https://ai.meta.com/blog/meta-LLaMA-3/}} This suggests that the amount of multilingual data in the pre-training phase may enhance the model's ability to generate more translatable rewrites. However, this advantage does not extend when comparing the \textsc{LLaMA} models to \textsc{Tower-Instruct}. Despite being inherently multilingual primarily trained on translation-related tasks, \textsc{Tower-Instruct} performs lower than the \textsc{LLaMA} models in translatability. This discrepancy can be attributed to \textsc{Tower-Instruct} not being specifically trained on rewriting tasks to improve MT quality, highlighting the importance of introducing translation-related knowledge for effective rewriting.

We further compare the results with off-the-shelf paraphrasing (\textsc{DIPPER}) and text-editing (\textsc{CoEdIT-XL}) tools. Despite being specifically trained for rewriting tasks, their rewrites are not as translatable as those generated by the prompted LLMs. For \textsc{DIPPER}, this may be due to its primary focus on paraphrasing, which has been shown to be less effective (\S \ref{simplification best}). In the case of \textsc{CoEdIT}, we attribute the lower performance to the model's smaller size (3B) compared to the 7B LLMs used for prompting.

\subsection{Same LLM vs. Different LLM}
\label{appendix:same llm}
We distinguish whether the LLM being prompted is the same as the one used as the MT system. Initially, we expected the highest improvements when prompting \textsc{Tower-Instruct}, which may incur self-preference bias, where the LLM favors its own outputs due to recognition \citep{panickssery2024llm}. However, our results indicate that prompting \textsc{Tower-Instruct} does not yield the most translatable rewrites. Instead, the LLaMA series models consistently outperform in this aspect. Interestingly, \textsc{Tower-Instruct} consistently produces rewrites that are more meaning-preserving compared to \textsc{LLaMA-2} or \textsc{LLaMA-3}, resulting in higher \textsc{xCOMET}$(s,t,r)$ scores overall. We conclude that prompting the same LLM used for the MT system is not helpful in generating more translatable rewrites, but these rewrites are better at preserving the intended meaning.

\section{Qualitative Evaluation}
\label{appendix: qualitative eval details}
\subsection{Copying Behavior}
\label{appendix:direct_copy}
To prevent LLMs from directly copying the original source, we explicitly state in the prompt to ``\textit{avoid directly copying the source}'' (Appendix \ref{appendix:prompt_templates}). However, we still observe some rewrites that are identical to the source sentence. We count the occurrences and compute the percentage per language pair in Table \ref{tab:direct_copy}. Note that we do not consider Translatability-Aware Selection rewrite method here since this involves selecting whether to keep the original source or use the rewrite based on translatability scores. The highest occurrence appears for stylistic rewrites using the \textsc{CoEdIT-XL} Coherent prompt, where the source is copied most of the time (82.2\%, 91.9\%, 93.2\% for \textsc{En-De}, \textsc{En-Ru}, and \textsc{En-Zh}, respectively).

\begin{figure}
    \centering
    \includegraphics[width=\linewidth]{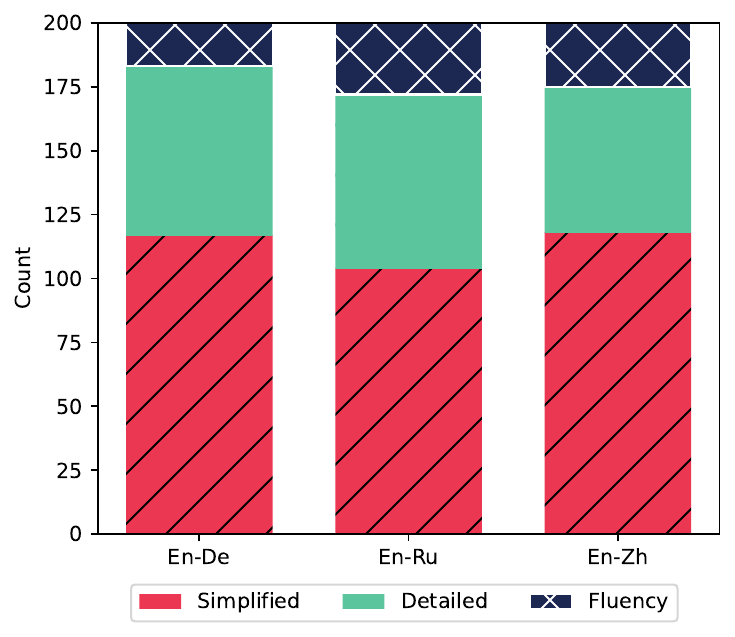}
    \caption{Distribution of properties of good rewrites.}
    \label{fig:success}
\end{figure}

\subsection{What makes a Good Rewrite for MT?}
Qualitatively examining translation outputs reveals several common patterns, which motivate us to conduct a detailed qualitative analysis. Here, we aim to identify the properties that lead to meaning-equivalent rewrites that are easier to translate. We examine 200 data instances where each rewrite is the highest performing rewrite based on the \textsc{xCOMET}$(s,t)$ score. To focus on successful rewrites, we filter instances where \textsc{xCOMET}$(s',t')$ $>$ \textsc{xCOMET}$(s,t)$. Each rewrite is annotated with the following labels: (1) \textbf{Simplified}: Replaces complex words with simpler ones or reduces structural complexity; (2) \textbf{Detailed}: Adds information for better context; (3) \textbf{Fluency}: Restructures the sentence for better flow and readability. 
Examples of rewrites for each annotation label are in  Table \ref{tab:success_types}. 

As shown in Figure \ref{fig:success}, most successful rewrites are labeled as \textbf{Simplified}. This highlights the effectiveness of simplification, which has been consistently effective even in the context of LLMs. Notably, many simplified rewrites involve changing complex words to simpler, more conventional alternatives (e.g., ``Derry City \textit{emerged victorious} in the President's Cup as they \textit{ran out} 2-0 \textit{winners} over Shamrock Rovers.'' → ``Derry City \textit{won} the President's Cup title by \textit{defeating} Shamrock Rovers 2-0.''). This finding aligns with our conclusions from MT-Agnostic rewriting methods (\S \ref{3.1 mt-agnostic}), where simplification emerged as the best rewrite method among the prompting variations.

\section{Additional Results}
\subsection{Additional LLM Baselines}
\label{appendix:more_llms}
\paragraph{LLMs for Rewriting.}
Our initial experiments consist of 21 input rewriting methods across 3 LLMs (\textsc{LLaMA-2 7B}, \textsc{LLaMA-3 8B}, and \textsc{Tower-Instruct 7B}). In Table \ref{tab:more_rewrite_llms}, we present extended experiment results by applying simplification rewriting with two additional LLMs: \textsc{Aya-23 8B} \citep{aryabumi2024aya23openweight} and \textsc{Tower-Instruct 13B} \citep{alves2024tower}. The results confirms that simplification rewriting improves translation quality measured by \textsc{xCOMET}$(s,t,r)$ compared to the original baseline.

\paragraph{LLMs for MT.}
Furthermore, we initially relied on \textsc{Tower-Instruct 7B} as our MT system for all our experiments since it is specifically trained for translation-related tasks and has demonstrated superior MT performance (\S \ref{3 method}). However, we extend our analysis by comparing the original baseline and our winning strategy (simplification with \textsc{Tower-Instruct 7B}) using two additional LLMs as the MT system. As shown in Table \ref{tab:more_mt_llms}, our method outperforms the original baseline in terms of both the translation quality (\textsc{xCOMET}$(s,t,r)$) and \textsc{MetricX}$(s,t)$, regardless of the LLM used as the MT system.

\subsection{Additional Language Pairs}
\label{appendix:more_lang_pairs}
To assess the generalizability to other source languages, we test two of our winning strategies (simplification with \textsc{Tower-Instruct 7B} and inference-time selection) on seven additional into-English and non-English language pairs from the WMT-23 General MT task test set.\footnote{\url{https://www2.statmt.org/wmt23/translation-task.html}} As shown in Table \ref{tab:more_lang_pairs}, while translatability scores (\textsc{xCOMET}$(s,t)$) improve across all language pairs, translation quality (\textsc{xCOMET}$(s,t,r)$) improvements are less pronounced compared to out-of-English pairs. Notably, gains in translation quality are observed only for German-English (\textsc{De-En}) and Chinese-English (\textsc{Zh-En}) pairs. These results highlight the importance of input rewrites' quality, which is currently higher for high-resource source languages. This motivates further work to strengthen input rewriting for broader range of source languages.

\section{Human Annotation Details}
\label{appendix:annotation details}
We use Qualtrics\footnote{\url{https://www.qualtrics.com}} to design our survey and Prolific\footnote{\url{https://www.prolific.com}} to recruit human annotators fluent in the tested target language.

\subsection{Original MT vs. Rewrite MT Details}
\label{appendix:mt_details}
We randomize the order of the two sentences (original MT and rewrite MT) to mitigate position bias. Annotators evaluate which sentence is better across four dimensions: fluency, understandability, level of detail, and meaning preservation. The entire survey is estimated to take approximately 20 minutes to complete. We recruit a total of 9 annotators and provide a compensation of 5 US dollars per survey (15 US dollars/hr), totaling 45 US dollars.

\subsection{Original vs. Rewrite Details}
\label{appendix:details}
Each annotator is tasked to judge how well the rewritten sentence preserves the meaning of the original source sentence. The survey is estimated to take approximately 30 minutes to complete. We recruit a total of 3 annotators. We offer a compensation of 7.5 US dollars per survey (15 US dollars/hr), totaling 22.5 US dollars.

\subsection{Annotator Instructions}
\begin{figure*}
    \centering
        \fbox{
        \includegraphics[width=350pt]{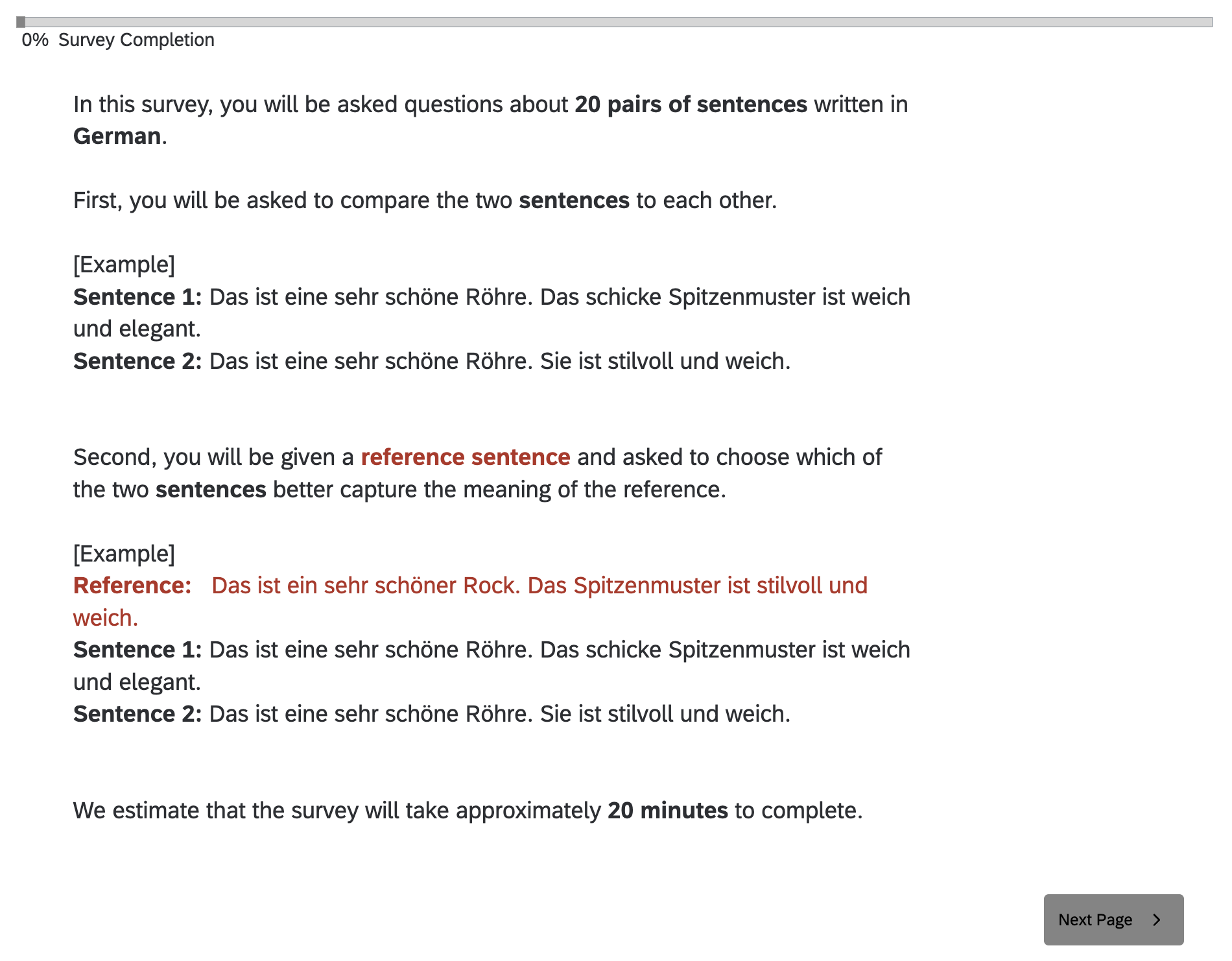}
    }
    \caption{Instructions to human annotators for Original MT vs. Rewrite MT evaluation.}
    \label{fig:human_intro}
\end{figure*}

\begin{figure*}
    \centering
        \fbox{
        \includegraphics[width=350pt]{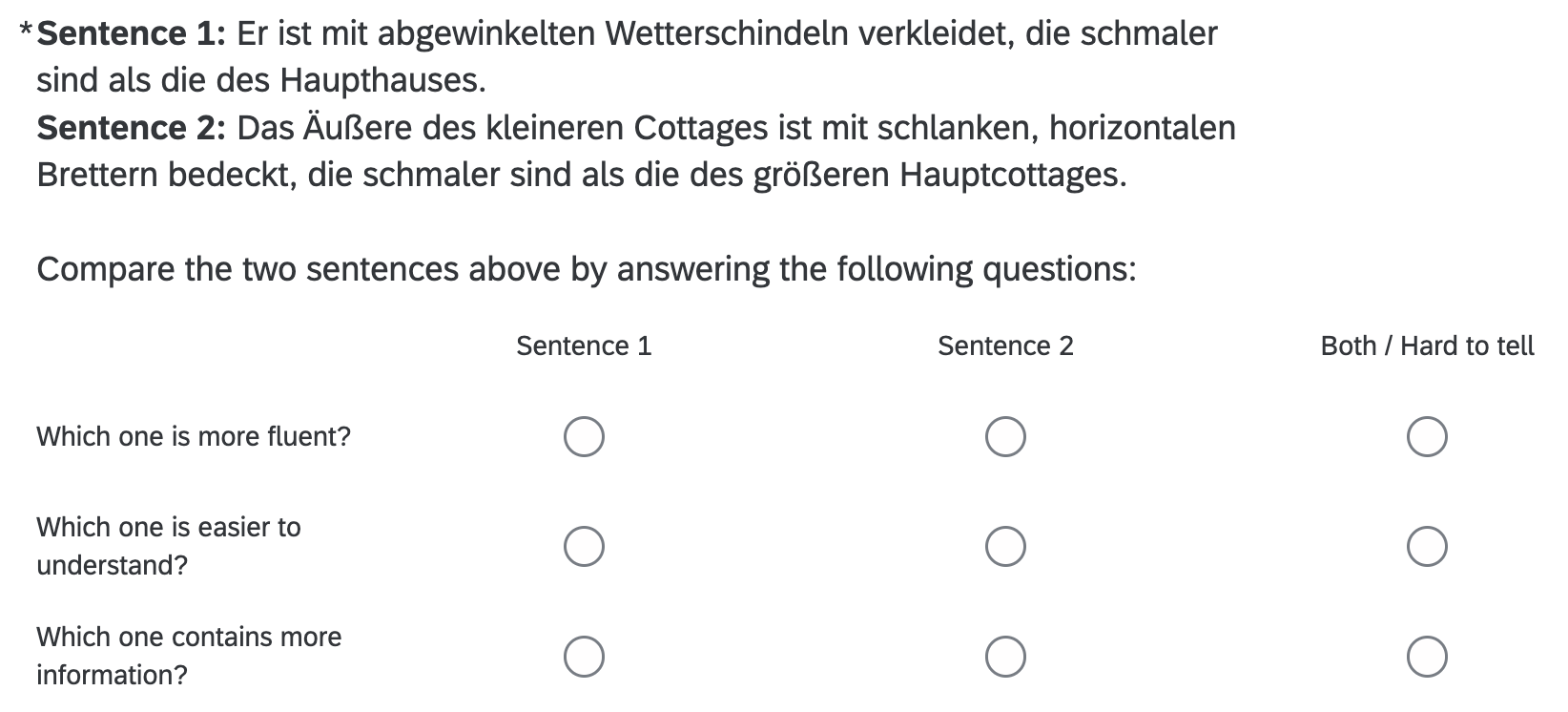}
    }
    \caption{Survey content of the first part to compare Original MT vs. Rewrite MT. To avoid position bias, we randomly shuffle the order of original translations ($t$) and translations of rewrites ($t'$) for \textbf{Sentence 1} and \textbf{2}.}
    \label{fig:human_example_1}
\end{figure*}

\begin{figure*}
    \centering
        \fbox{
        \includegraphics[width=350pt]{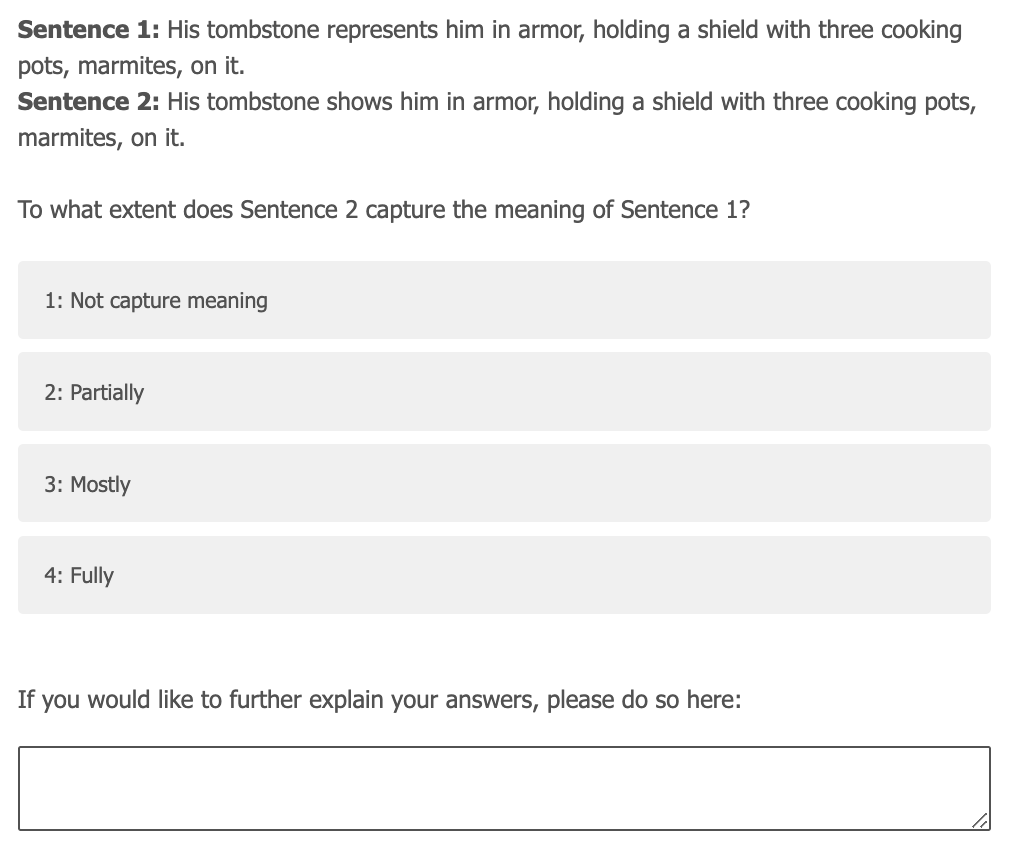}
    }
    \caption{Survey content to compare Original (\textbf{Sentence 1}) vs. Rewrite (\textbf{Sentence 2}).}
    \label{fig:human_example_3}
\end{figure*}

In Figures \ref{fig:human_intro} to \ref{fig:human_example_3}, we present the instructions and survey content provided to annotators. For the Original MT vs. Rewrite MT evaluation, each annotator reviews 20 sets of examples. Each question consists of two parts: \textbf{1)} comparing the two sentences based on fluency, understandability, and level of detail, and \textbf{2)} selecting which sentence better preserves the meaning of the reference translation. For the Original vs. Rewrite evaluation, each annotator reviews 30 sets of examples. Additionally, a free-form text box is provided alongside each example for annotators to offer feedback or suggestions.

\section{Time \& Computational Efficiency}
\label{appendix:inference_cost}
We show that on average, rewriting with our winning strategy is not a resource-intensive option for downstream applications in terms of both time and computation. For approximately 1.5K sentences, the rewrite and MT pipeline using our winning strategy (simplification with \textsc{Tower-Instruct 7b} takes 1 hour, compared to 30 minutes for the MT process alone. All variants of our prompting experiments are conducted using a single NVIDIA RTX958 A5000 GPU. In terms of efficiency compared to automatic post-editing (\S \ref{res:post-editing}), both approaches remains equivalent in time and computational requirements since the rewriting or post-editing process only differs in its position within the pipeline. Input rewriting modifies the source before the MT system, while output post-editing adjusts the translation after the MT system.


\begin{table*}[!htp]
\centering
\resizebox{\textwidth}{!}{%
    \begin{tabular}{l l}
    \specialrule{1.3pt}{0pt}{0pt}
    \textbf{Rewrite} & \textbf{Prompt} \\ \midrule
    
    \multirow{4}{*}{\textbf{Simplification}} & Simplify the English sentence. Simplification may include identifying complex words and replacing with simpler \\
    & or shorter words or using active voice instead of passive voice. Try to keep the meaning of the Original sentence. \\
    & Original: \textit{This is a very nice skirt. The lacy pattern is classy and soft.} \\
    & Simplified: \\ \midrule

    \multirow{3}{*}{\textbf{Paraphrase}} & Paraphrase the English sentence. Try to not directly copy but keep the meaning of the Original sentence. \\
    & Original: \textit{This is a very nice skirt. The lacy pattern is classy and soft.} \\
    & Paraphrase: \\ \midrule

    \multirow{4}{*}{\textbf{Stylistic} (\textsc{CoEdIT})} & \textbf{(GEC)} Fix the grammar: \\
    & \textbf{(Coherent)} Make this text coherent: \\
    & \textbf{(Understandable)} Rewrite to make this easier to understand: \\
    & (\textbf{Formal)} Write this more formally: \\ \midrule

    \multirow{3}{*}{\textbf{Easy Translation}} & Rewrite the Original sentence to be easier for translation in {target language}. New sentence should be in English. \\
    & Original: \textit{This is a very nice skirt. The lacy pattern is classy and soft.} \\
    & New: \\ \midrule

    \multirow{8}{*}{\textbf{CoT}} & \textbf{(Step 1)} Rewrite the Original English sentence to New English sentence that translates better into German. \\
    & Avoid directly copying the Original sentence while keeping its meaning. New sentence should be in English. \\
    & Original: \textit{This is a very nice skirt. The lacy pattern is classy and soft.} \\
    & New: \\
    
    \\
    & \textbf{(Step 2)} Now, translate the English sentence to German. \\
    & English: \\
    & German: \\
    
    \specialrule{1.3pt}{0pt}{0pt}
    \end{tabular}
}
\caption{Exemplar prompt templates for English-German language pair used for prompting experiments. \textit{Italic} represents the source sentence used in this example.}
\label{tab:prompting_template}
\end{table*}
\definecolor{light gray}{rgb}{0.898, 0.902, 0.906}
\definecolor{light pink}{rgb}{0.996, 0.9102, 0.9219}

\begin{table*}[htbp]
\centering
\resizebox{\textwidth}{!}
{
    \begin{tabular}{l}
    \specialrule{1.3pt}{0pt}{0pt}
    \textbf{Basic} \\ \midrule
    
    {\textsc{\#\#\#} \textbf{Instruction:}} Rewrite this English sentence to give a better translation. {\texttt{\textbackslash n\textbackslash n}} \\
        
    {\textsc{\#\#\#} \textbf{English}:} This is a very nice skirt. The lacy pattern is classy and soft.{\texttt{\textbackslash n}} \\
    
    {\textsc{\#\#\#} \textbf{English rewrite}:} The lacy pattern on this skirt is elegant and soft. \\
    \midrule

    \textbf{MT} \\ \midrule
    
    {\textsc{\#\#\#} \textbf{Instruction:}} Rewrite this English sentence to give a better translation in German. German sentence is the hypothesis translation that \\
    we are trying to improve.{\texttt{\textbackslash n\textbackslash n}} \\
    
    {\textsc{\#\#\#} \textbf{English}:} This is a very nice skirt. The lacy pattern is classy and soft.{\texttt{\textbackslash n}} \\

    {\textsc{\#\#\#} \textbf{German}:} Das ist eine sehr schöne Röhre. Das schicke Spitzenmuster ist weich und elegant.{\texttt{\textbackslash n}} \\
    
    {\textsc{\#\#\#} \textbf{English rewrite}:} The lacy pattern on this skirt is elegant and soft. \\
    \midrule

    \textbf{Reference} \\ \midrule
    
    {\textsc{\#\#\#} \textbf{Instruction:}} Rewrite this English sentence to give a better translation in German. German sentence is the human-annotated translation \\
    that we are trying to pursue.{\texttt{\textbackslash n\textbackslash n}} \\
    
    {\textsc{\#\#\#} \textbf{English}:} This is a very nice skirt. The lacy pattern is classy and soft.{\texttt{\textbackslash n}} \\

    {\textsc{\#\#\#} \textbf{German}:} Das ist ein sehr schöner Rock. Das Spitzenmuster ist stilvoll und weich.{\texttt{\textbackslash n}} \\
    
    {\textsc{\#\#\#} \textbf{English rewrite}:} The lacy pattern on this skirt is elegant and soft. \\
    \specialrule{1.3pt}{0pt}{0pt}
    \end{tabular}
}
\caption{Exemplar prompt templates for supervised fine-tuning experiments (English-German pair). We additionally give machine translation for the \textbf{MT} prompt and reference translation for the \textbf{Reference} prompt after \textsc{\#\#\#} \textbf{German:}.}
\label{tab:training_template}
\end{table*}

\clearpage

\begin{table*}[!htp]
\centering
\resizebox{400pt}{!}{%
    \begin{tabular}{l l l}
    \specialrule{1.3pt}{0pt}{0pt}
    \textbf{Split} & \textbf{Dataset} & \textbf{\# Sentences}  \\ \midrule
    
    \multirow{2}{*}{\textbf{Train}} & Source and positive rewrite pairs for SFT (English-German, $\mathcal{D}_{pos}$) & 7,016 \\
    & Source and positive rewrite pairs for SFT (English-Russian, $\mathcal{D}_{pos}$) & 8,126 \\ \midrule
    
    \multirow{6}{*}{\textbf{Test}} & WMT-23 General MT Task (English-German) & 1,557 \\
    & WMT-23 General MT Task (English-Russian) & 2,074 \\
     & WMT-23 General MT Task (English-Chinese) & 3,074 \\
 & WMT-23 General MT Task (English-Czech) & 2,074 \\
    & WMT-23 General MT Task (English-Hebrew) & 2,074 \\
     & WMT-23 General MT Task (English-Japanese) & 2,074 \\
    
    \specialrule{1.3pt}{0pt}{0pt}
    \end{tabular}
}
\caption{Summary statistics of training and test datasets.}
\label{tab:dataset_details}
\end{table*}
\definecolor{light blue}{RGB}{215, 242, 252}
\definecolor{light purple}{RGB}{247, 215, 252}
\definecolor{light orange}{rgb}{0.9961, 0.875, 0.7188}

\begin{table*}[!htp]
\centering
\resizebox{\textwidth}{!}{%
\renewcommand{\arraystretch}{1.1}
\begin{tabular}{c l l l l l l l}
    \specialrule{1.3pt}{0pt}{0pt}
    \textbf{Language} & \textbf{Type} & \textbf{Prompt/Model} & \textbf{\textsc{xCOMET}$(s,t)$} & \textbf{\textsc{xCOMET}$(s,r)$} & \textbf{\textsc{xCOMET}$(s,t,r)$} & \textbf{\textsc{MetricX}$(s,t)$} & \textbf{\textsc{MetricX}$(t,r)$} \\ \midrule
    
    \multirow{22}{*}{\large \textbf{\textsc{en-de}}} & \textbf{Original} & - & 0.893 & 0.904 & 0.898 & 2.038 & 1.534 \\
    \cmidrule(lr){2-8}
    
    & \fcolorbox{white}{light blue}{\raisebox{-0.2em}{\includegraphics[height=1em]{figures/logos/agnostic.png}} \textbf{MT-Agnostic}} & \textbf{Simplification} (\small \textsc{LLaMA-2}) & \textbf{0.931} & 0.846 & 0.900 & \textbf{1.185} & 1.727 \\
    & &(\small \textsc{LLaMA-3}) & \textbf{0.944} & 0.820 & 0.903 & \textbf{0.925*} & 1.600 \\
    & &(\small \textsc{Tower-Instruct}) & \textbf{0.922} & 0.885 & \textbf{0.907} & 1.504 & 1.519 \\
    
    & & \textbf{Paraphrase} (\small \textsc{LLaMA-2}) & \textbf{0.926} & 0.823 & 0.889 & \textbf{1.126} & \textbf{1.480} \\
    & & (\small \textsc{LLaMA-3}) & \textbf{0.938} & 0.796 & 0.892 & \textbf{0.955} & \textbf{1.469} \\
    & & (\small \textsc{Tower-Instruct}) & 0.902 & 0.887 & 0.901 & \textbf{1.310} & 1.534 \\

    & &(\small \textsc{DIPPER} (L80/O60)) & \textbf{0.904} & 0.745 & 0.838 & \textbf{1.674} & 2.757 \\
    & &(\small \textsc{DIPPER} (L80/O40)) & \textbf{0.913} & 0.797 & 0.863 & \textbf{1.461} & 2.266 \\
    & &(\small \textsc{DIPPER} (L60/O40)) & \textbf{0.917} & 0.847 & 0.892 & \textbf{1.555} & 1.958 \\

    & &\textbf{Stylistic} (\small \textsc{CoEdIT} GEC) & \textbf{0.901} & 0.899 & 0.900 & \textbf{1.709} & 1.555 \\
    & & (\small \textsc{CoEdIT} Coherent) & 0.898 & 0.900 & 0.898 & \textbf{1.728} & 1.595 \\
    & & (\small \textsc{CoEdIT} Understandable) & \textbf{0.949} & \textbf{0.758} & 0.862 & \textbf{0.989} & 2.610 \\
    & & (\small \textsc{CoEdIT} Formal) & \textbf{0.937} & 0.830 & 0.900 & \textbf{1.063} & 1.879 \\

    \cmidrule(lr){2-8}
    & \fcolorbox{white}{light purple}{\raisebox{-0.2em}{\includegraphics[height=1em]{figures/logos/task.png}} \textbf{Task-Aware}} & \textbf{Easy Translation} (\small \textsc{LLaMA-2}) & \textbf{0.916} & 0.857 & 0.893 & \textbf{1.654} & 2.482 \\
    & & (\small \textsc{LLaMA-3}) & \textbf{0.932} & 0.827 & 0.899 & \textbf{1.151} & 2.241 \\
    & & (\small \textsc{Tower-Instruct}) & \textbf{0.901} & 0.900 & 0.903 & \textbf{1.759} & 2.427 \\
    & & \textbf{CoT} (\small \textsc{Tower-Instruct}) & \textbf{0.907} & 0.816 & 0.897 & \textbf{1.892} & 1.578 \\
    
    \cmidrule(lr){2-8}
    & \fcolorbox{white}{light orange}{\raisebox{-0.2em}{\includegraphics[height=1em]{figures/logos/translatability.png}} \textbf{Translatability-Aware}} & \textbf{Selection} & \textbf{0.921} & 0.907 & \textbf{0.915*} & \textbf{1.734} & \textbf{1.461*} \\
    & & \textbf{Fine-tune} (\small Basic) & \textbf{0.934} & 0.851 & \textbf{0.909} & \textbf{1.878} & \textbf{1.499} \\
    & & (\small MT) & \textbf{0.919} & 0.856 & 0.903 & \textbf{1.947} & 1.593 \\
    & & (\small Reference) & 0.896 & 0.836 & 0.876 & 2.023 & 2.028 \\
    

    \specialrule{1.3pt}{0pt}{0pt}
    \end{tabular}
}
\caption{Detailed results of English-German pair using different rewrite methods. Statistically significant average improvements ($p$-value $< 0.05$) are \textbf{bold}. Best scores for each metric is \textbf{bold} with \textbf{*}. \textsc{xCOMET}$(s,t)$: translatability (↑); \textsc{xCOMET}$(s,r)$: meaning preservation (↑); \textsc{xCOMET}$(s,t,r)$: overall translation quality (↑); \textsc{MetricX}$(s,t)$: quality estimation (↓); \textsc{MetricX}$(t,r)$: reference-based metric (↓). For \textsc{DIPPER} \cite{dipper} variations, L and O denote lexical and order diversity, respectively.}
\label{tab:detailed_results_ende}
\end{table*}
\clearpage

\definecolor{light blue}{RGB}{215, 242, 252}
\definecolor{light purple}{RGB}{247, 215, 252}
\definecolor{light orange}{rgb}{0.9961, 0.875, 0.7188}

\begin{table*}[!htp]
\centering
\resizebox{\textwidth}{!}{%
\renewcommand{\arraystretch}{1.1}
\begin{tabular}{c l l l l l l l}
    \specialrule{1.3pt}{0pt}{0pt}
    \textbf{Language} & \textbf{Type} & \textbf{Prompt/Model} & \textbf{\textsc{xCOMET}$(s,t)$} & \textbf{\textsc{xCOMET}$(s,r)$} & \textbf{\textsc{xCOMET}$(s,t,r)$} & \textbf{\textsc{MetricX}$(s,t)$} & \textbf{\textsc{MetricX}$(t,r)$} \\ \midrule

    \multirow{22}{*}{\large \textbf{\textsc{en-ru}}} & \textbf{Original} & - & 0.872 & 0.884 & 0.868 & 2.535 & 2.028 \\
    \cmidrule(lr){2-8}

    & \fcolorbox{white}{light blue}{\raisebox{-0.2em}{\includegraphics[height=1em]{figures/logos/agnostic.png}} \textbf{MT-Agnostic}} & \textbf{Simplification} (\small \textsc{LLaMA-2}) & \textbf{0.916} & 0.839 & \textbf{0.882} & \textbf{0.951} & 2.160 \\
    & &(\small \textsc{LLaMA-3}) & \textbf{0.919} & 0.812 & \textbf{0.885} & \textbf{0.804} & 2.039 \\
    & &(\small \textsc{Tower-Instruct}) & \textbf{0.921} & 0.870 & \textbf{0.891} & \textbf{1.135} & \textbf{1.921} \\

    & & \textbf{Paraphrase} (\small \textsc{LLaMA-2}) & \textbf{0.923} & 0.804 & \textbf{0.881} & \textbf{0.882} & \textbf{1.853} \\
    & &  (\small \textsc{LLaMA-3}) & \textbf{0.930} & 0.788 & \textbf{0.882} & \textbf{0.855} & \textbf{1.863} \\
    & & (\small \textsc{Tower-Instruct}) & \textbf{0.887} & 0.878 & \textbf{0.878} & \textbf{1.095} & \textbf{1.976} \\

    & & (\small \textsc{DIPPER} (L80/O60)) & \textbf{0.904} & 0.735 & 0.821 & \textbf{1.249} & 3.476 \\
    & & (\small \textsc{DIPPER} (L80/O40)) & \textbf{0.909} & 0.790 & 0.853 & \textbf{1.105} & 2.773 \\
    & & (\small \textsc{DIPPER} (L60/O40)) & \textbf{0.905} & 0.834 & 0.873 & \textbf{1.119} & 2.418 \\

    & & \textbf{Stylistic} (\small \textsc{CoEdIT} GEC) & 0.873 & 0.884 & 0.869 & \textbf{1.327} & \textbf{1.969} \\
    & & (\small \textsc{CoEdIT} Coherent) & 0.873 & 0.884 & 0.869 & \textbf{1.368} & \textbf{1.989} \\
    & & (\small \textsc{CoEdIT} Understandable) & \textbf{0.918} & 0.801 & 0.873 & \textbf{0.991} & 2.726 \\
    & & (\small \textsc{CoEdIT} Formal) & \textbf{0.916} & 0.841 & \textbf{0.887} & \textbf{0.922} & 2.020 \\
    \cmidrule(lr){2-8}

    & \fcolorbox{white}{light purple}{\raisebox{-0.2em}{\includegraphics[height=1em]{figures/logos/task.png}} \textbf{Task-Aware}} & \textbf{Easy Translation} (\small \textsc{LLaMA-2}) & \textbf{0.914} & 0.839 & \textbf{0.884} & \textbf{1.037} & 10.849 \\
    & & (\small \textsc{LLaMA-3}) & \textbf{0.917} & 0.808 & \textbf{0.881} & \textbf{0.801*} & 10.401 \\
    & & (\small \textsc{Tower-Instruct}) & \textbf{0.885} & 0.883 & \textbf{0.878} & \textbf{1.277} & 11.137 \\
    
    & & \textbf{CoT} (\small \textsc{Tower-Instruct}) & \textbf{0.903} & 0.871 & 0.875 & \textbf{2.432} & 2.024 \\
    \cmidrule(lr){2-8}

    & \fcolorbox{white}{light orange}{\raisebox{-0.2em}{\includegraphics[height=1em]{figures/logos/translatability.png}} \textbf{Translatability-Aware}} & \textbf{Selection} & \textbf{0.914} & \textbf{0.891*} & \textbf{0.899*} & \textbf{2.096} & \textbf{1.830*} \\
    
    & & \textbf{Fine-tune} (\small Basic) & \textbf{0.912} & 0.848 & \textbf{0.886} & \textbf{2.123} & \textbf{1.932} \\
    & & (\small MT) & \textbf{0.904} & 0.851 & 0.871 & \textbf{2.119} & \textbf{1.997} \\
    & & (\small Reference) & \textbf{0.881} & 0.812 & 0.859 & \textbf{2.284} & 2.012 \\
    

    \specialrule{1.3pt}{0pt}{0pt}
    \end{tabular}
}
\caption{Detailed results of English-Russian pair using different rewrite methods.}
\label{tab:detailed_results_enru}
\end{table*}
\definecolor{light blue}{RGB}{215, 242, 252}
\definecolor{light purple}{RGB}{247, 215, 252}
\definecolor{light orange}{rgb}{0.9961, 0.875, 0.7188}

\begin{table*}[!htp]
\centering
\resizebox{\textwidth}{!}{%
\renewcommand{\arraystretch}{1.1}
\begin{tabular}{c l l l l l l l}
    \specialrule{1.3pt}{0pt}{0pt}
    \textbf{Language} & \textbf{Type} & \textbf{Prompt/Model} & \textbf{\textsc{xCOMET}$(s,t)$} & \textbf{\textsc{xCOMET}$(s,r)$} & \textbf{\textsc{xCOMET}$(s,t,r)$} & \textbf{\textsc{MetricX}$(s,t)$} & \textbf{\textsc{MetricX}$(t,r)$} \\ \midrule

    \multirow{19}{*}{\large \textbf{\textsc{en-zh}}} & \textbf{Original} & - & 0.786 & 0.775 & 0.794 & 3.445 & 2.282 \\

    \cmidrule(lr){2-8}

    & \fcolorbox{white}{light blue}{\raisebox{-0.2em}{\includegraphics[height=1em]{figures/logos/agnostic.png}} \textbf{MT-Agnostic}} & \textbf{Simplification} (\small \textsc{LLaMA-2}) & \textbf{0.828} & 0.728 & 0.796 & \textbf{1.321} & 2.537 \\
    &  & (\small \textsc{LLaMA-3}) & \textbf{0.823} & 0.701 & 0.795 & \textbf{1.252*} & 2.572 \\
    & & (\small \textsc{Tower-Instruct}) & \textbf{0.821} & 0.759 & \textbf{0.802} & \textbf{1.521} & \textbf{2.227} \\

    & & \textbf{Paraphrase} (\small \textsc{LLaMA-2}) & \textbf{0.818} & 0.683 & 0.771 & \textbf{1.330} & 2.478 \\
    & & (\small \textsc{LLaMA-3}) & \textbf{0.826} & 0.662 & 0.766 & \textbf{1.341} & 2.534 \\
    & & (\small \textsc{Tower-Instruct}) & \textbf{0.797} & 0.765 & 0.798 & \textbf{1.580} & 2.283 \\

    & & (\small \textsc{DIPPER} (L80/O60)) & \textbf{0.813} & 0.622 & 0.722 & \textbf{1.583} & 4.009 \\
    & &  (\small \textsc{DIPPER} (L80/O40)) & \textbf{0.816} & 0.670 & 0.750 & \textbf{1.499} & 3.196 \\
    & & (\small \textsc{DIPPER} (L60/O40)) & \textbf{0.809} & 0.711 & 0.775 & \textbf{1.503} & 2.725\\
    
    & & \textbf{Stylistic} (\small \textsc{CoEdIT} GEC) & 0.789 & 0.772 & 0.795 & \textbf{1.632} & 2.251 \\
    & &  (\small \textsc{CoEdIT} Coherent) & 0.786 & 0.774 & 0.794 & \textbf{1.658} & 2.267 \\
    & &  (\small \textsc{CoEdIT} Understandable) & \textbf{0.839*} & 0.677 & 0.774 & \textbf{1.358} & 3.174 \\
    & &  (\small \textsc{CoEdIT} Formal) & \textbf{0.823} & 0.730 & 0.798 & \textbf{1.336} & 2.443 \\
    \cmidrule(lr){2-8}

    & \fcolorbox{white}{light purple}{\raisebox{-0.2em}{\includegraphics[height=1em]{figures/logos/task.png}} \textbf{Task-Aware}} & \textbf{Easy Translation} (\small \textsc{LLaMA-2}) & \textbf{0.821} & 0.721 & 0.784 & \textbf{1.900} & 7.732 \\
    & & (\small \textsc{LLaMA-3}) & \textbf{0.830} & 0.687 & 0.783 & \textbf{1.360} & 7.608 \\
    & & (\small \textsc{Tower-Instruct}) & \textbf{0.793} & 0.762 & 0.791 & \textbf{1.618} & 7.650 \\
    
    & & \textbf{CoT} (\small \textsc{Tower-Instruct}) & \textbf{0.821} & 0.769 & 0.771 & \textbf{3.321} & 2.432 \\

    \cmidrule(lr){2-8}
    & \fcolorbox{white}{light orange}{\raisebox{-0.2em}{\includegraphics[height=1em]{figures/logos/translatability.png}} \textbf{Translatability-Aware}} & \textbf{Selection} & \textbf{0.823} & \textbf{0.783*} & \textbf{0.819*} & \textbf{3.149} & \textbf{2.206*} \\

    \specialrule{1.3pt}{0pt}{0pt}
    \end{tabular}
}
\caption{Detailed results of English-Chinese pair using different rewrite methods.}
\label{tab:detailed_results_enzh}
\end{table*}
\clearpage

\definecolor{light blue}{RGB}{215, 242, 252}
\definecolor{light purple}{RGB}{247, 215, 252}
\definecolor{light orange}{rgb}{0.9961, 0.875, 0.7188}

\begin{table*}[!htp]
\centering
\resizebox{330pt}{!}{%
\begin{tabular}{l l l l l l}
    \specialrule{1.3pt}{0pt}{0pt}
    \textbf{Type} & \textbf{Prompt/Model} & \textbf{\textsc{en-de}} & \textbf{\textsc{en-ru}} & \textbf{\textsc{en-zh}} \\ \midrule
    
    \fcolorbox{white}{light blue}{\textbf{MT-Agnostic}} & \textbf{Simplification} (\small \textsc{LLaMA-2}) & 2.06	& 2.37	& 2.37 \\
    & (\small \textsc{LLaMA-3}) & 0.39	& 0.33	& 0.29 \\
    & (\small \textsc{Tower-Instruct}) & 28 &	29.3	& 30.2 \\

    & \textbf{Paraphrase} (\small \textsc{LLaMA-2}) & 0 & 0 & 0 \\
    &  (\small \textsc{LLaMA-3}) & 0.06	&0.07	&0.03 \\
    & (\small \textsc{Tower-Instruct}) & 37.3	& 38.2	& 38 \\

    & (\small \textsc{DIPPER} (L80/O60)) & 0.19	& 0.94	& 1.04 \\
    & (\small \textsc{DIPPER} (L80/O40)) & 0.51& 1.5 &	1.53 \\
    & (\small \textsc{DIPPER} (L60/O40)) & 1.48	& 2.5	& 2.44 \\

    & \textbf{Stylistic} (\small \textsc{CoEdIT} GEC) & 42.6& 44 & 48.3 \\
    & (\small \textsc{CoEdIT} Coherent) & \color{red}{82.2}	& \color{red}{91.9}	& \color{red}{93.2} \\
    & (\small \textsc{CoEdIT} Understandable) & 1.61& 1.88	& 1.53 \\
    & (\small \textsc{CoEdIT} Formal) & 5.33& 3.76	& 5.5 \\
    \midrule

    \fcolorbox{white}{light purple}{\textbf{Task-Aware}} & \textbf{Easy Translation} (\small \textsc{LLaMA-2}) & 3.04 & 3.55 & 3.63 \\
    & (\small \textsc{LLaMA-3}) & 0.24 & 0.66 & 0.27 \\
    & (\small \textsc{Tower-Instruct}) & 12.3 & 18.6 & 15.4 \\
    & \textbf{CoT} (\small \textsc{Tower-Instruct}) & 0.71& 1.45& 1.53 \\
    \midrule

    \fcolorbox{white}{light orange}{\textbf{Translatability-Aware}} & \textbf{Fine-tune} (\small Basic) & 4.5	& 3.91 & - \\
    & (\small MT) &3.73	& 3.42 & - \\
    & (\small Reference) & 6.17	& 7.85 & - \\


    \specialrule{1.3pt}{0pt}{0pt}
    \end{tabular}
}
\caption{Percentage of occurrence (\%) where the rewrite is a direct copy of the original source sentence.}
\label{tab:direct_copy}
\end{table*}



\begin{CJK*}{UTF8}{gbsn}

\begin{table*}[!htp]
\centering
\resizebox{\textwidth}{!}{%
\renewcommand{\arraystretch}{1.1}
\begin{tabular}{l p{4cm} p{4cm} p{4cm} p{4cm} l l}
    \specialrule{1.3pt}{0pt}{0pt}
    \textbf{Label} & \textbf{Original} & \textbf{Rewrite} & \textbf{Original MT}  & \textbf{Rewrite MT} & \textbf{\textsc{xCOMET}$(s,t)$} & \textbf{\textsc{xCOMET}$(s',t')$} \\ \midrule

    \textbf{Simplified} & When Michael ``Hopper'' McGrath \textbf{lobbed} a ball in, Molloy \textbf{leapt} highest before rifling a sublime goal to the roof of the net. & When Michael McGrath \textbf{threw} the ball in, Molloy \textbf{jumped} highest and scored a beautiful goal to the top of the net. & Als Michael ``Hopper'' McGrath einen Ball hereinwarf, sprang Molloy am höchsten und schoss einen herrlichen Treffer auf das Dach des Netzes. & Als Michael McGrath den Ball in die Luft warf, sprang Molloy am höchsten und erzielte einen wunderschönen Treffer in die obere Netzhöhe. & 0.906 & 0.945 \\
        
    \cmidrule(lr){2-7}

    & Derry City \textbf{emerged victorious} in the President's Cup as they ran out 2-0 winners over Shamrock Rovers. & Derry City \textbf{won} the President's Cup title by defeating Shamrock Rovers 2-0. & Derry City 在总统杯赛中获胜，以 2-0 的比分击败尚洛克罗弗斯。& Derry City 以 2-0 的 比分击败 Shamrock Rovers，获得了总统杯冠军。& 0.648 & 0.952 \\
    \midrule

    \textbf{Detailed} & The great majority of rankers never advanced beyond principalis. & The vast majority of soldiers remained in the lowest rank throughout their careers. & Die große Mehrheit der Reiter schaffte es nie über den Rang eines principalis. & Die überwiegende Mehrheit der Soldaten blieb während ihrer gesamten Karriere in der niedrigsten Ränge. & 0.938 & 0.982 \\
    \cmidrule(lr){2-7}

    & I've noticed you almost need line of sight for it to work. & It appears that visibility plays a crucial role in the effectiveness of the process. & \russian{Я заметил, что для работы вам почти все время нужен прямой свет.} & \russian{Похоже, что видимость играет решающую роль в эффективности процесса.} & 0.98 & 1.0 \\
    \midrule

    \textbf{Fluency} & It's a thing I've never said before either. & I've never said that before either. & \russian{Это то, что я никогда не говорил раньше.} & \russian{Я никогда этого не говорил и раньше.} & 0.989 & 1.0 \\
    \cmidrule(lr){2-7}

    & When I started in summer with those multi-source experiments. & I began a series of experiments in the summer. & 我在夏天开始进行多来源实验时。& 我在夏天开始了一系列的实验。& 0.858 & 1.0 \\

    \specialrule{1.3pt}{0pt}{0pt}
    \end{tabular}
}
\caption{Examples of rewrites for each annotation label (\textbf{Simplified}, \textbf{Detailed} and \textbf{Fluency}).}
\label{tab:success_types}
\end{table*}

\end{CJK*}

\clearpage

\begin{table*}[!htp]
\centering
\resizebox{\textwidth}{!}{%
\renewcommand{\arraystretch}{1.1}
\begin{tabular}{p{3.5cm} p{4cm} p{4cm} p{4cm} p{4cm} l l l l}
    \specialrule{1.3pt}{0pt}{0pt}
    \textbf{Prompt/Model} & \textbf{Original} & \textbf{Rewrite} & \textbf{Original MT}  & \textbf{Rewrite MT} & \textbf{Flesch$(s)$} & \textbf{Flesch$(s')$} & \textbf{WSTF$(t)$} & \textbf{WSTF$(t')$} \\ \midrule

    \textbf{Simplification} (\textsc{LLaMA-3}) & She \textbf{steamed via} Hawaii, Midway, Guam, and Subic Bay for Vietnam and anchored in the Saigon River on 13 September. & She \textbf{sailed from} Hawaii to Vietnam, stopping at Midway, Guam, and Subic Bay, and \textbf{arrived at} the Saigon River on September 13. & Sie fuhr über Hawaii, Midway, Guam und Subic Bay nach Vietnam und ankerte am 13. September in der Saigon-Schifffahrt. & Sie segelte von Hawaii nach Vietnam, machte Halt in Midway, Guam und Subic Bay und erreichte am 13. September in der Saigon River. & 74.53 & \textbf{76.56} & 1.032 & \textbf{0.838} \\ \midrule

    \textbf{Simplification} (\textsc{Tower-Instruct}) & The remnants of Felix continued northeastward across the Atlantic until dissipating near Shetland on August 25. & Felix's remnants continued northeastward across the Atlantic until dissipating near Shetland on August 25. & Die Überreste von Felix zogen sich über den Atlantik in nordöstlicher Richtung bis zum 25. August, als sie sich in der Nähe von Shetland auflösten. & Felix's Reste zogen sich über den Atlantik in nordöstlicher Richtung bis zum 25. August, als sie sich in der Nähe von Shetland auflösten. & 31.89 & \textbf{38.32} & 1.193 & \textbf{1.109} \\ 
    \cmidrule{2-9}

     & Cambrai thus \textbf{reverted}, but only briefly, to the Western Frankish Realm. & Cambrai \textbf{returned} to the Western Frankish Realm, but only briefly. & Cambrai fiel daher, aber nur kurzzeitig, wieder an das Westfrankenreich zurück. & Cambrai kehrte zum Westfrankenreich zurück, aber nur kurz. & \textbf{68.77} & 54.22 & 0.728 & \textbf{0.429} \\
    
    \specialrule{1.3pt}{0pt}{0pt}
    \end{tabular}
}
\caption{Examples of simplification rewrites for English-German (\textsc{En-De}) pair and their corresponding input and output readability scores. \textbf{Flesch}: Flesch Reading Ease score (↑); \textbf{WSTF}: Vienna formula (↓).}
\label{tab:readability_ende}
\end{table*}

\begin{table*}[!htp]
\centering
\resizebox{\textwidth}{!}{%
\renewcommand{\arraystretch}{1.1}
\begin{tabular}{p{3.5cm} p{4cm} p{4cm} p{4cm} p{4cm} l l l l}
    \specialrule{1.3pt}{0pt}{0pt}
    \textbf{Prompt/Model} & \textbf{Original} & \textbf{Rewrite} & \textbf{Original MT}  & \textbf{Rewrite MT} & \textbf{Flesch$(s)$} & \textbf{Flesch$(s')$} & \textbf{Flesch-Ru$(t)$} & \textbf{Flesch-Ru$(t')$} \\ \midrule

    \textbf{Simplification} (\textsc{LLaMA-3}) & Later, Wallachia's Vornic Radu Socol traveled to Suceava, bringing Despot two steeds, a kuka hat with precious stones, and 24,000 ducats. & Radu Socol, the Vornic of Wallachia, visited Suceava and brought two horses, a hat with precious stones, and 24,000 ducats to Despot. & \russian{Позже, Ворник Раду Соколь из Валахии отправился в Сучаву, привезнув деспоту двух лошадей, кукушку с драгоценными камнями и 24 000 дукатов.} & \russian{Раду Сокол, вонник Валахии, посетил Сучаву и принес деспоту два коня, шляпу с драгоценными камнями и 24 000 дукатов.} & \textbf{67.08} & 66.07 & 55.81 & \textbf{64.80} \\ \midrule

    \textbf{Simplification} (\textsc{Tower-Instruct}) & \textbf{Appalled} at the thought of Emily \textbf{cavorting} with Casey, Margo \textbf{vindictively revealed} Emily's \textbf{hooker past} to Tom and Casey. & Margo was shocked that Emily was hanging out with Casey and so she told Tom and Casey about Emily's past as a prostitute. & \russian{Потрясенная мыслью о том, что Эмили развлекается с Кейси, Марго мстительно рассказала Тому и Кейси о прошлом Эмили проституткой.} & \russian{Марго была потрясена тем, что Эмили общалась с Кейси, и поэтому она рассказала Тому и Кейси о прошлом Эмили как проститутке.} & 60.65 & \textbf{81.97} & 58.47 & \textbf{64.40} \\
    
    \specialrule{1.3pt}{0pt}{0pt}
    \end{tabular}
}
\caption{Examples of simplification rewrites for English-Russian (\textsc{En-Ru}) pair and their corresponding input and output readability scores. \textbf{Flesch}: Flesch Reading Ease score (↑); \textbf{Flesch-Ru}: Russian version of Flesch (↑).}
\label{tab:readability_enru}
\end{table*}

\begin{CJK*}{UTF8}{gbsn}

\begin{table*}[!htp]
\centering
\resizebox{\textwidth}{!}{%
\renewcommand{\arraystretch}{1.1}
\begin{tabular}{p{3.5cm} p{4cm} p{4cm} p{4cm} p{4cm} l l l l}
    \specialrule{1.3pt}{0pt}{0pt}
    \textbf{Prompt/Model} & \textbf{Original} & \textbf{Rewrite} & \textbf{Original MT}  & \textbf{Rewrite MT} & \textbf{Flesch$(s)$} & \textbf{Flesch$(s')$} \\ \midrule

    \textbf{Simplification} (\textsc{LLaMA-3}) & During the delay, the tire carcass wrapped itself around the axle, costing him several laps. & The tire wrapped around the axle, causing him to lose several laps. & 延迟期间，轮胎壳破损，裹住了轮毂，让他失去了几圈的速度。& 轮胎缠在轴上，让他失去了几圈。& 64.71 & \textbf{84.68} \\ \midrule

    \textbf{Simplification} (\textsc{Tower-Instruct}) & Japanese artillery attempted to engage them but South Dakota and the other battleships easily outranged them. & Japanese artillery tried to attack them but South Dakota and the other battleships were too far away. & 日本炮兵试图与他们交战，但南达科他和其他战舰的射程远远超过他们。& 日本炮兵试图袭击他们，但南达科他和其他战舰太远了。 & 38.32 & \textbf{62.68} \\

    \specialrule{1.3pt}{0pt}{0pt}
    \end{tabular}
}
\caption{Examples of simplification rewrites for English-Chinese (\textsc{En-Zh}) pair and their corresponding input readability scores. \textbf{Flesch}: Flesch Reading Ease score (↑).}
\label{tab:readability_enzh}
\end{table*}

\end{CJK*}
\clearpage

\begin{table*}[!htp]
\centering
\resizebox{\textwidth}{!}{%
\renewcommand{\arraystretch}{1.1}
\begin{tabular}{c l l l l l l}
    \specialrule{1.3pt}{0pt}{0pt}
    \textbf{Language} & \textbf{Prompt/Model} & \textbf{\textsc{xCOMET}$(s,t)$} & \textbf{\textsc{xCOMET}$(s,t,r)$} & \textbf{\textsc{MetricX}$(s,t)$} & \textbf{\textsc{MetricX}$(t,r)$} \\ \midrule
    
    \multirow{3}{*}{\large \textbf{\textsc{en-de}}} & Original & 0.893 & 0.898 & 2.038 & 1.534 \\
    & Simplification (\small \textsc{Aya-23 8B}) & 0.901 & 0.900 & 1.956 & 1.624 \\
    & Simplification (\small \textsc{Tower-Instruct 13B}) & \textbf{0.924} & \textbf{0.912} & \textbf{1.562} & \textbf{1.445} \\
    \midrule

    \multirow{3}{*}{\large \textbf{\textsc{en-ru}}} & Original & 0.872 & 0.868 & 2.535 & 2.028 \\
    & Simplification (\small \textsc{Aya-23 8B}) & 0.880 & \textbf{0.875} & 2.428 & 1.938 \\
    & Simplification (\small \textsc{Tower-Instruct 13B}) & \textbf{0.901} & \textbf{0.889} & \textbf{2.137} & \textbf{1.861} \\

    \bottomrule
    \end{tabular}
}
\caption{Results with two additional LLMs for rewriting: \textsc{Aya-23 8B} and \textsc{Tower-Instruct 13B}. Statistically significant average improvements ($p$-value $< 0.05$) are \textbf{bold}. \textsc{xCOMET}$(s,t)$: translatability (↑); \textsc{xCOMET}$(s,t,r)$: overall translation quality (↑); \textsc{MetricX}$(s,t)$: quality estimation (↓); \textsc{MetricX}$(t,r)$: reference-based metric (↓).}
\label{tab:more_rewrite_llms}
\end{table*}
\begin{table*}[!htp]
\centering
\resizebox{\textwidth}{!}{%
\renewcommand{\arraystretch}{1.1}
\begin{tabular}{c l l l l l l l}
    \specialrule{1.3pt}{0pt}{0pt}
    \textbf{Language} & \textbf{MT System} & \textbf{Prompt/Model} & \textbf{\textsc{xCOMET}$(s,t)$} & \textbf{\textsc{xCOMET}$(s,t,r)$} & \textbf{\textsc{MetricX}$(s,t)$} & \textbf{\textsc{MetricX}$(t,r)$} \\ \midrule
    
    \multirow{6}{*}{\large \textbf{\textsc{en-de}}} & \multirow{2}{*}{\textsc{Tower-Instruct 7B}} & Original & 0.893 & 0.898 & 2.038 & 1.534 \\
    & & Simplification & \textbf{0.915} & \textbf{0.907} & 1.504 & 1.519 \\
    \cmidrule{2-7}
    
    & \multirow{2}{*}{\textsc{Aya-23 8B}} & Original & 0.887 & 0.891 & 1.926 & 1.554 \\
    & & Simplification & \textbf{0.911} & \textbf{0.902} & 1.660 & 1.571 \\
    \cmidrule{2-7}
    
    & \multirow{2}{*}{\textsc{Tower-Instruct 13B}} & Original & 0.880 & 0.887 & 2.043 & 1.522 \\
    & & Simplification & \textbf{0.900} & \textbf{0.893} & \textbf{1.778} & 1.556 \\
    \midrule

    \multirow{6}{*}{\large \textbf{\textsc{en-ru}}} & \multirow{2}{*}{\textsc{Tower-Instruct 7B}} & Original & 0.872 & 0.868 & 2.535 & 2.028 \\
    & & Simplification & \textbf{0.921} & \textbf{0.891} & \textbf{1.135} & \textbf{1.921} \\
    \cmidrule{2-7}
    
    & \multirow{2}{*}{\textsc{Aya-23 8B}} & Original & 0.863 & 0.852 & 2.711 & 2.323 \\
    & & Simplification & \textbf{0.892} & \textbf{0.872} & \textbf{2.300} & \textbf{2.173} \\
    \cmidrule{2-7}
    
    & \multirow{2}{*}{\textsc{Tower-Instruct 13B}} & Original & 0.887 & 0.882 & 2.290 & 1.915 \\
    & & Simplification & \textbf{0.894} & 0.875 & 2.296 & 1.915 \\
    \midrule

    \multirow{6}{*}{\large \textbf{\textsc{en-zh}}} & \multirow{2}{*}{\textsc{Tower-Instruct 7B}} & Original & 0.786 & 0.794 & 3.445 & 2.282 \\
    & & Simplification & \textbf{0.821} & \textbf{0.802} & \textbf{1.521} & \textbf{2.227} \\
    \cmidrule{2-7}
    
    & \multirow{2}{*}{\textsc{Aya-23 8B}} & Original & 0.769 & 0.779 & 3.758 & 2.572 \\
    & & Simplification & \textbf{0.793} & \textbf{0.788} & \textbf{3.433} & 2.530 \\
    \cmidrule{2-7}
    
    & \multirow{2}{*}{\textsc{Tower-Instruct 13B}} & Original & 0.755 & 0.764 & 3.421 & 2.341 \\
    & & Simplification & \textbf{0.772} & 0.767 & 3.236 & 2.413 \\

    \specialrule{1.3pt}{0pt}{0pt}
    \end{tabular}
}
\caption{Results with two additional LLMs as MT system: \textsc{Aya-23 8B} and \textsc{Tower-Instruct 13B}. Simplification is done with \textsc{Tower-Instruct 7b}. Statistically significant average improvements ($p$-value $< 0.05$) over their respective original baselines are \textbf{bold}. \textsc{xCOMET}$(s,t)$: translatability (↑); \textsc{xCOMET}$(s,t,r)$: overall translation quality (↑); \textsc{MetricX}$(s,t)$: quality estimation (↓); \textsc{MetricX}$(t,r)$: reference-based metric (↓).}
\label{tab:more_mt_llms}
\end{table*}
\begin{table*}[!htp]
\centering
\resizebox{\textwidth}{!}{%
\renewcommand{\arraystretch}{1.1}
\begin{tabular}{c l l l l l l l}
    \specialrule{1.3pt}{0pt}{0pt}
    \textbf{Language} & \textbf{Prompt/Model} & \textbf{\textsc{xCOMET}$(s,t)$} & \textbf{\textsc{xCOMET}$(s,t,r)$} & \textbf{\textsc{MetricX}$(s,t)$} & \textbf{\textsc{MetricX}$(t,r)$} \\ \midrule
    
    \multirow{3}{*}{\large \textbf{\textsc{cs-uk}}} & Original & 0.866 & 0.755 & 2.437 & 4.033 \\
    & Simplification & \textbf{0.885} & 0.749 & 2.355 & 4.053 \\
    & Selection & \textbf{0.930} & 0.748 & 3.050 & 4.053 \\
    \midrule

    \multirow{3}{*}{\large \textbf{\textsc{de-en}}} & Original & 0.969 & 0.622 & 1.869 & 4.760 \\
    & Simplification & \textbf{0.975} & \textbf{0.632} & 1.856 & 4.600 \\
    & Selection & \textbf{0.979} & \textbf{0.631} & 1.856 & 4.599 \\
    \midrule

    \multirow{3}{*}{\large \textbf{\textsc{he-en}}} & Original & 0.582 & 0.556 & 8.057 & 5.758 \\
    & Simplification & 0.562 & 0.514 & 8.671 & 6.374 \\
    & Selection & \textbf{0.639} & 0.514 & 9.192 & 6.541 \\
    \midrule

    \multirow{3}{*}{\large \textbf{\textsc{ja-en}}} & Original & 0.884 & 0.841 & 3.473 & 2.688 \\
    & Simplification & 0.896 & 0.828 & 3.303 & 2.929 \\
    & Selection & \textbf{0.918} & 0.827 & 3.659 & 2.964 \\
    \midrule

    \multirow{3}{*}{\large \textbf{\textsc{ru-en}}} & Original & 0.938 & 0.921 & 3.024 & 1.823 \\
    & Simplification & 0.945 & 0.922 & 2.909 & 1.879 \\
    & Selection & \textbf{0.954} & 0.923 & 3.079 & 1.912 \\
    \midrule

    \multirow{3}{*}{\large \textbf{\textsc{uk-en}}} & Original & 0.934 & 0.929 & 2.959 & 1.507 \\
    & Simplification & \textbf{0.951} & 0.929 & 2.684 & 1.595 \\
    & Selection & \textbf{0.962} & 0.929 & 3.055 & 1.656 \\
    \midrule

    \multirow{3}{*}{\large \textbf{\textsc{zh-en}}} & Original & 0.797 & 0.524 & 5.099 & 5.666 \\
    & Simplification & \textbf{0.809} & \textbf{0.530} & 4.849 & 5.582 \\
    & Selection & \textbf{0.827} & 0.528 & 5.202 & 5.800 \\

    \bottomrule
    \end{tabular}
}
\caption{Results with into-English and non-English language pairs. Simplification is done with \textsc{Tower-Instruct 7b}. Statistically significant average improvements ($p$-value $< 0.05$) over their respective original baselines are \textbf{bold}. \textsc{xCOMET}$(s,t)$: translatability (↑); \textsc{xCOMET}$(s,t,r)$: overall translation quality (↑); \textsc{MetricX}$(s,t)$: quality estimation (↓); \textsc{MetricX}$(t,r)$: reference-based metric (↓).}
\label{tab:more_lang_pairs}
\end{table*}

\definecolor{maryred}{rgb}{0.758, 0.109, 0.0234}

\begin{figure*}
    \centering
        \fbox{
        \includegraphics[width=350pt]{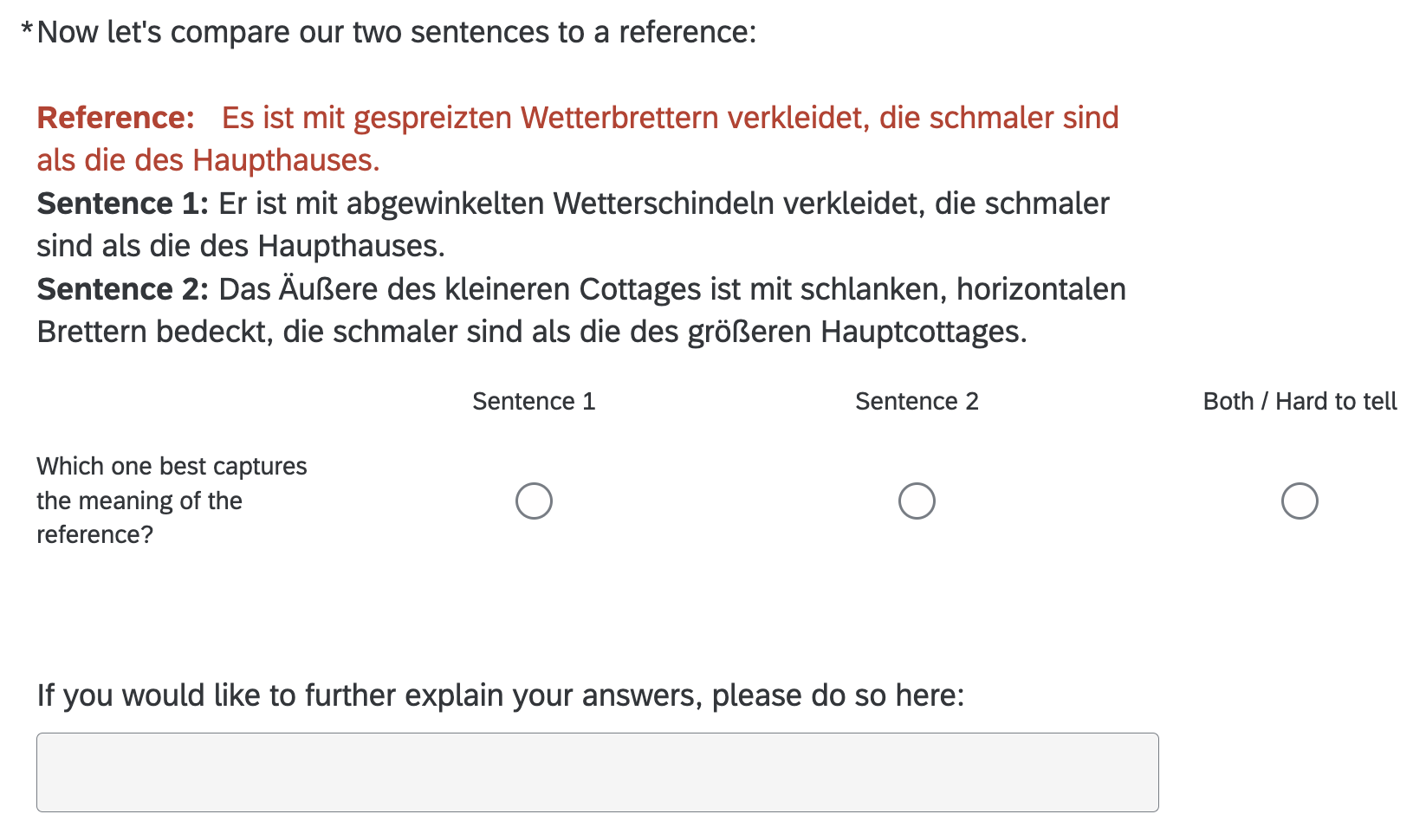}
    }
    \caption{Survey content of the second part to compare to the \color{maryred}{\textbf{Reference translation}}\color{black}{. An optional text box is given for each example for further comments.}}
    \label{fig:human_example_2}
\end{figure*}

\end{document}